\documentclass[lettersize,journal]{IEEEtran}

\usepackage{amsmath,amsfonts}

\usepackage{algorithmic}
\usepackage{algorithm}
\usepackage{array}
\usepackage[caption=false,font=normalsize,labelfont=sf,textfont=sf]{subfig}
\usepackage{textcomp}
\usepackage{stfloats}
\usepackage{url}
\usepackage{verbatim}
\usepackage{graphicx}
\usepackage{cite}
\usepackage{float}
\usepackage{gensymb}
\usepackage{multirow}

\usepackage{xcolor,colortbl}
\usepackage{amsmath, xparse}

\usepackage{bm}
\usepackage{mathtools}
\usepackage{commath}
\usepackage{color}
\usepackage{soul}
\usepackage{amssymb}
\usepackage{comment}
\usepackage{xfrac}
\usepackage{nicefrac}
\usepackage{hyperref}

\usepackage[
singlelinecheck=false 
]{caption}

\begin{document}

\title{AntGrip - Boosting Parallel Plate Gripper Performance Inspired by the Internal Hairs of Ant Mandibles}

\author{Mohamed Sorour, and Barbara Webb
\thanks{Insect Robotics Group - Institute for Perception, Action and Behaviour,
School of Informatics, University of Edinburgh. Informatics Forum, 10 Crichton St, EH8 9AB Edinburgh, United Kingdom.
        {\tt\small msorour@ed.ac.uk}}%
}



\maketitle
\begin{abstract}
Ants use their mandibles - effectively a two-finger gripper - for a wide range of grasping activities. Here we investigate whether mimicking the internal hairs found on ant mandibles can improve performance of a two-finger parallel plate robot gripper. With bin picking applications in mind, the gripper fingers are long and slim, with interchangeable soft gripping pads that can be hairy or hairless. A total of $2400$ video-documented experiments have been conducted, comparing hairless to hairy pads with different hair patterns. Simply by adding hairs, the grasp success rate was increased by at least $29\%$, and the number of objects that remain securely gripped during manipulation more than doubled. This result not only advances the state of the art in grasping technology, but also provides novel insight into the mechanical role of mandible hairs in ant biology.
\end{abstract}


\section{Introduction}
It is increasingly recognised that practical grasping scenarios, such as warehouse automation, require a combination of friction- and suction-based grasping \cite{Correll2016,Fujita2020,Sorour2019}. Although the latter was most successful in the recent Amazon Picking Challenge, and have become a widely used technology, they are not sufficient on their own for a successful grasping system. For example, cylindrical and parallelpiped objects are better grasped using two or three fingered grippers \cite{Chen2015}. In applications, therefore, suction cups are commonly teamed up with parallel plate grippers \cite{Yu2016,Gustavo2020,Majumder2020,Morrison2017} (a variation of the two fingered grippers), or other form of friction grasping \cite{Fujita2019,Hernandez2017} unless severe object-related assumptions are presented \cite{Huang2022,Valencia2017,Mantriota2011,Avigal2022,Kang2019,Pham2019}. As a consequence, any improvement in the performance of two-finger friction based grippers could have significant impact, particularly if it is relatively low cost to implement.

\begin{figure}[t!]
\centering
\includegraphics[width=0.7\linewidth]{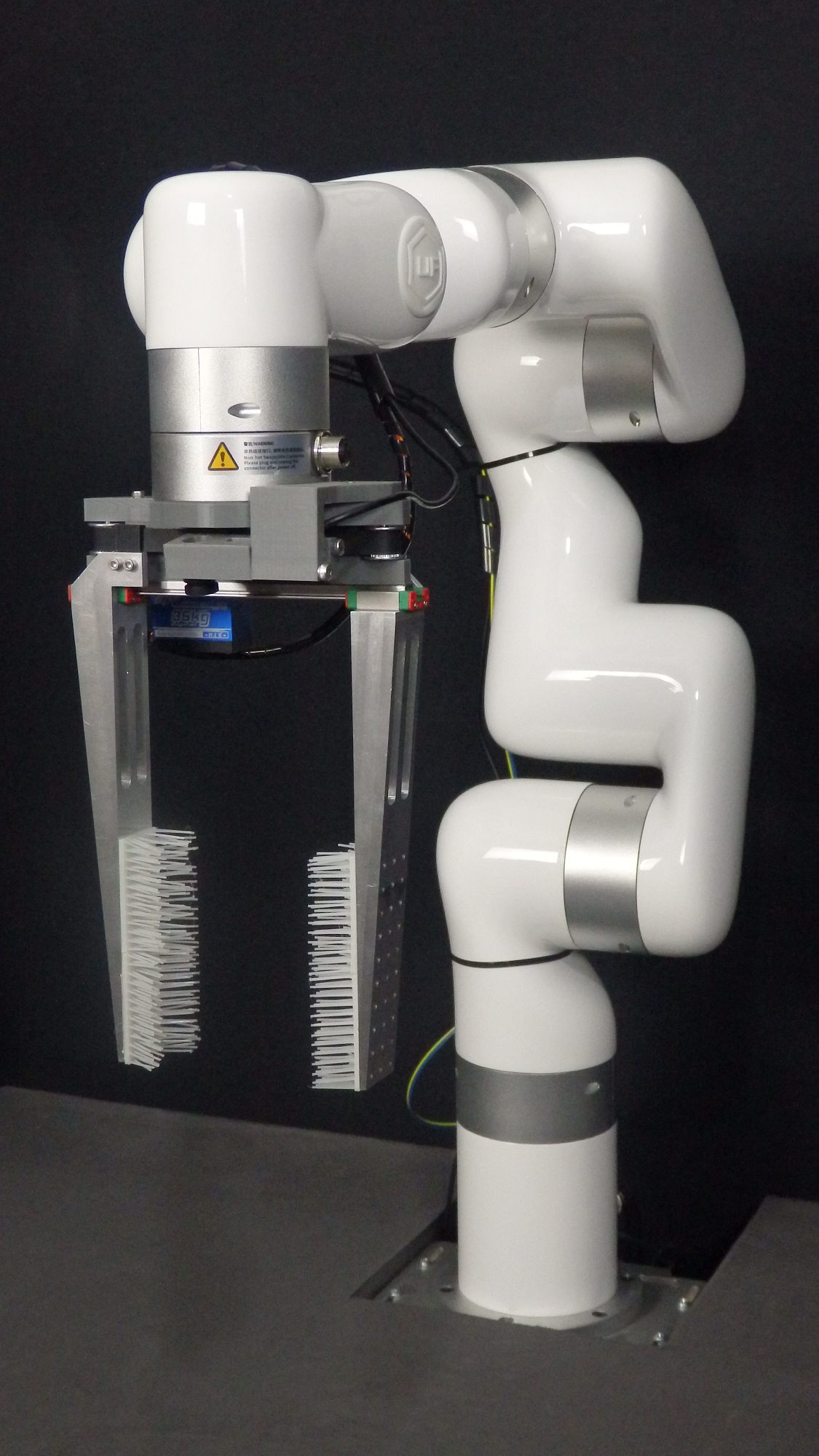}
\captionsetup{belowskip=-10pt}
\captionsetup{aboveskip=5pt}
\caption{Ant inspired gripper "AntGrip" fitted to the uFactory xArm$6$ robot manipulator.}
\label{figure_Antgrip_setup}
\end{figure}

The recognised need for multimodal grippers has let to the development of two fingered grippers with suction capability \cite{Kang2019, Nakamoto2018, WadeMcCue2017, Nechyporenko2021, Wang2019}; and three fingered designs with two under-actuated fingers and the third being a suction cup \cite{Hasegawa2017,Hasegawa2019}, tailored for warehouse automation. In \cite{Vu2020}, a four fingered gripper with separate suction is developed. Suction-assisted soft robot grippers have also attracted much attention due to being adaptive to the geometric shape of most household objects of interest, for example, an underactuated soft hand employing suction at finger tips is introduced in \cite{Wu2018}. Suction at finger tips is also developed in \cite{Yamaguchi2013} for grasping flat surfaces and rectangular objects. However the complexity and bulkiness of these designs has limited their adoption for real applications. 

\begin{figure*}[t!]
\centering
\includegraphics[width=1.0\textwidth]{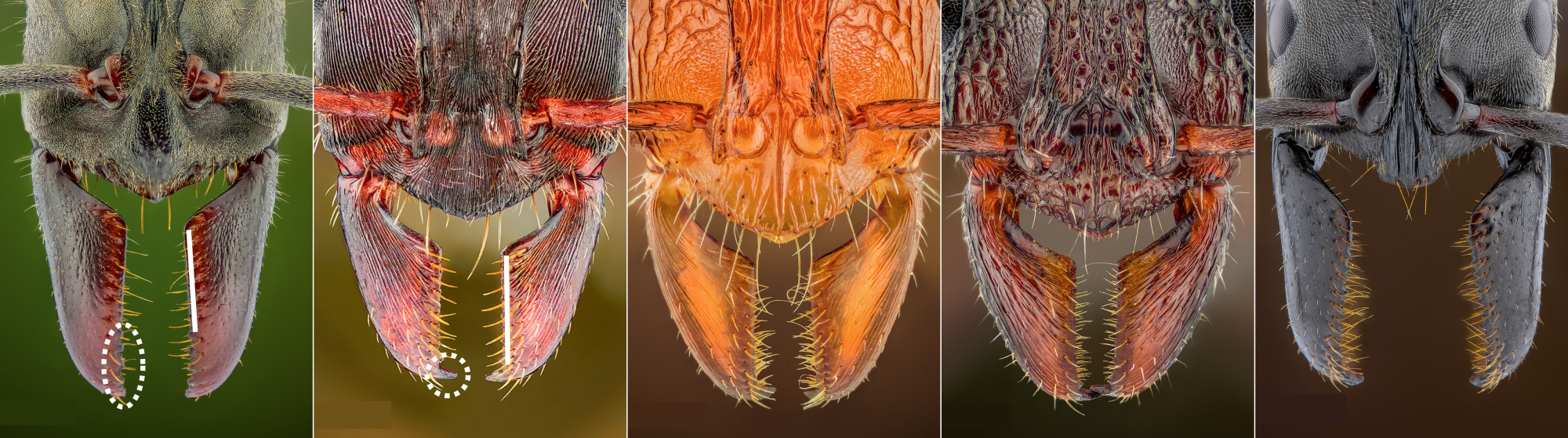}
\caption{Internal hairs of ant mandibles in \textit{ophthalmopone berthoudi}, \textit{ectatomma bruneum}, \textit{ectatomma tuberculatum}, \textit{rhytidoponera sp.}, and \textit{paltothyreus tarsatus} respectively from left to right, adapted from \cite{ant_photos}. Circles and lines in left panels highlight tooth-like and cutting structures respectively.}
\label{figure_internal_hair_of_ant_mandibles}
\end{figure*}

Among friction-based grippers, the two-fingered is the most common due to its simplicity, cost effectiveness and ease of control. The parallel plate variation is most convenient for tight space applications such as bin picking, shelf stowing, etc. However, two fingers have limitations in grasping cylindrical shaped objects due to having limited contact, often leading manufacturers to prefer three fingers \cite{RightPick} for better grasp stability despite being more bulky. 
Attempts to improve/customize parallel plate grippers are numerous. To name only a few, in \cite{Wang2021}, a scooping-binding variation is developed targeting the food industry, and in \cite{Kobayashi2019}, the gripper is designed to have a varying width depending on how widely it is opened, in an effort to make the gripper smaller in size.  In \cite{Guo2017, Harada2011}, compliant gripping tips are examined to increase the grasp robustness of parallel jaw grippers.


Here we take a novel approach to the problem of improving parallel plate grippers by seeking inspiration from the largest population of two-fingered gripper users, namely the ants. Ants use their mandibles, which consist of two rigid opposed jaws, each hinged, and actuated separately (Fig. \ref{figure_internal_hair_of_ant_mandibles}), for a huge range of grasping activities, from transporting delicate eggs to dismembering prey. As such, it seems plausible that examination of ant grasping could provide insights to improve the performance of two-finger robot grippers. Yet, despite recognition that "from the perspective of a roboticist, the variety and refinement of ant manipulation skills can
be startling and humbling" \cite{mason2018toward}, to the best of the authors' knowledge, the only work closely studying ants in the context of robotic grasping is introduced in \cite{Zhang2020}, where a model gripper is developed inspired by the \textit{harpegnathos venator} ants, in an attempt to replicate its ability to grasp delicate objects in narrow spaces. 

In this work, we focus on the role of the internal hairs of ant mandibles, shown in Fig. \ref{figure_internal_hair_of_ant_mandibles} for a selection of ant species. These hairs, described in detail in \cite{Gronenberg1994}, are divided into a set of hair fields at the base of the mandible and "ordinary" sensory hairs along the mandible. Shorter hairs along the internal mandible structure are described as mechanosensory hairs, and have been largely regarded as sensory elements \cite{Boublil2021} in the biological literature. There is evidence, however, that some ant hairs provide passive mechanical support for items carried by ants. The ammochaetae or psammophores, form a collection of long hairs in the lower head parts of some ant species, and have been shown to play a vital role in supporting water \cite{Wheeler1907} and sand \cite{Spangler1966} particles, as well as foraged wheat and small seeds \cite{Porter1990} grasped by mandibles during transportation. 

Drawing from these observations, we have developed a parallel plate gripper fitted with hairy pads to evaluate the impact of such hairs on grasp success/robustness as compared to a hairless counterpart. Two performance metrics are used to assess the results, the most conservative of which indicates a performance boost of at least $29\%$, and up to $40\%$ if the hair configuration is fine tuned. The contribution of the work presented in this paper is twofold:
\begin{itemize}
    \item Extending the capabilities of two fingered grippers, particularly to make robust grasps of cylindrical objects.
    \item Redefining the role of internal hairs of ant mandibles by demonstrating the potential impact of such hairs in securing a stable grasp.
\end{itemize}

The paper is organised as follows: section II presents the ant inspired gripper design. Experiment design methodology, as well as the results, interpretations and future work are reported in section III. Conclusions are finally given in section IV.

\begin{figure}[t!]
\centering
\includegraphics[width=1.0\linewidth]{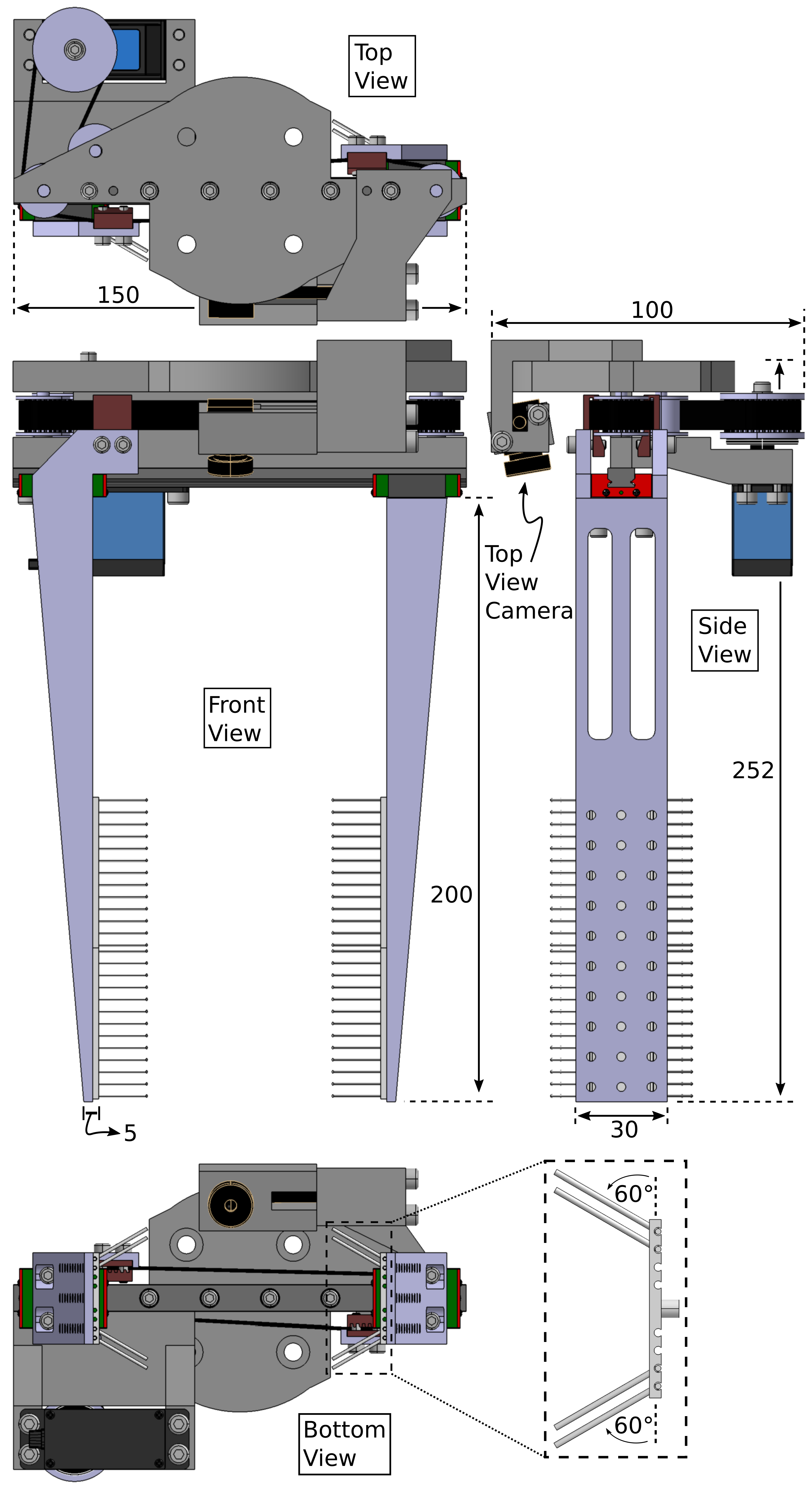}
\captionsetup{belowskip=-10pt}
\captionsetup{aboveskip=5pt}
\caption{Ant inspired gripper "AntGrip" CAD model in front, side, top, and bottom views, key dimensions shown in millimeter.}
\label{figure_gripper_CAD_model}
\end{figure}


\section{Ant Inspired Gripper}

General observation of the ant's mandible internal structure in Fig. \ref{figure_internal_hair_of_ant_mandibles} reveals three distinctive features. First (highlighted by circles in the left panels of Fig. \ref{figure_internal_hair_of_ant_mandibles}) the mandibles end with irregular out-protruding tooth-like structures (sometimes appearing sharp), to penetrate/pierce when biting prey or breaking items down into pieces. Second, marked by a white line in the respective panels, are long edges which together with a tooth-tip form an inverted scissors-like structure to perform the cutting process. Third, the internal hairs protruding at both sides of the mandible, seen from the top view in Fig \ref{figure_internal_hair_of_ant_mandibles}. Our hypothesis is that these hairs mechanically support grasped objects and help the ant to achieve stable grasps, especially for irregular shaped objects that form the majority of ant grasping activities. Note that although the morphology of the mandibles can vary substantially, often adapted to the task allocated per individual per ant species \cite{PAUL2001,Klunk2021,Gronenberg1997}, the internal hairs appear to be a common feature amongst many species.

In this work, we adopt the use of internal hairs without imitating other aspects of the ant mandibles, given that our target application differs in some critical ways from the natural domain of ants. Most targets for ant grasps are irregular, rough objects such as food items; whereas warehouse automation usually features smooth surfaces and
continuous geometric shapes due to industrial packaging. To this end, while a rotating two fingered gripper (mandibles) would serve the ant very well due to the necessity of a scissors-like cutting behaviour, it would fail to satisfy these robot grasping interests as compared to sliding fingers. In addition, bin/shelf picking/stowing applications require accessing objects from a distance due to tight spaces or a cluttered environment. This, as well as manipulation hardware safety, favors lengthy, slim grippers with inherent or active compliance. These guidelines are used as the main requirements for our application-oriented gripper design, enabling independent evaluation of the impact of internal hairs on grasp success/stability.

\begin{figure}[b!]
\centering
\includegraphics[width=1.0\linewidth]{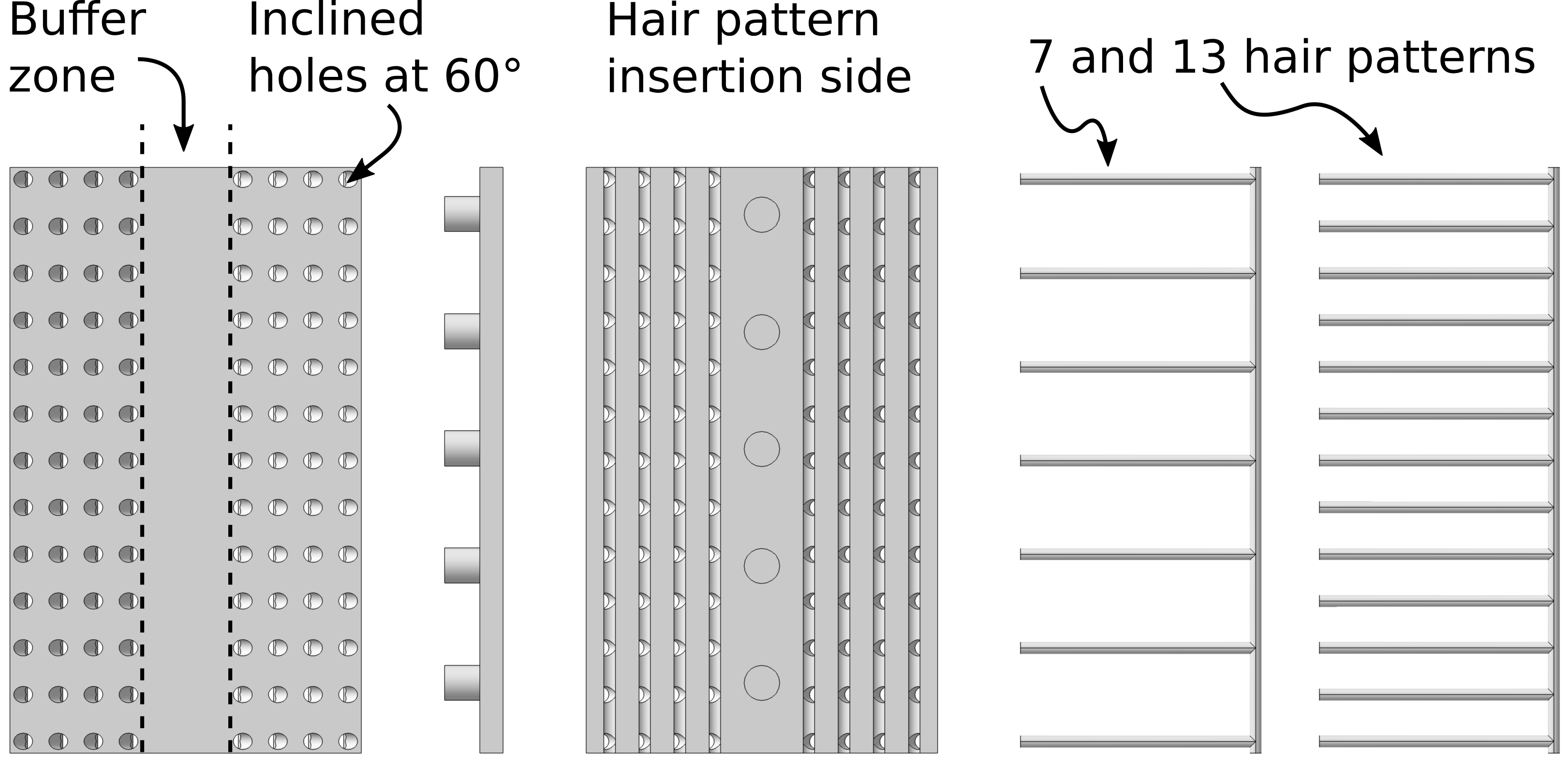}
\captionsetup{belowskip=-10pt}
\captionsetup{aboveskip=5pt}
\caption{$60\degree$ gripping pad (left) and the $2$ hair patterns (right) used in this work to construct the hairy pads.}
\label{figure_pad_and_hair_CAD_model}
\end{figure}

The CAD model of the designed gripper is shown in Fig. \ref{figure_gripper_CAD_model}. The fingers are wedge shaped, to allow them to move in between cluttered objects with minimal disturbance, long ($200mm$) and thin ($5mm$ thick fingertip including the hair pad) for the intended tight-space applications. Although the actuator can generate up to $200N$ of axial force, the maximum grasping force applicable is limited to $40N$ to avoid structural elastic deformation. Actuation power is transferred to the fingers via timing belt transmission, with each finger clamped to one side of the belt as depicted in the top view in Fig. \ref{figure_gripper_CAD_model}. This introduces a limited amount of passive compliance. Each finger is fixed to a linear bearing/rail system constraining its motion to the gripper's single DoF. The minimum and maximum gripper opening is $5mm$ and $110mm$ respectively, with positioning accuracy of $\pm{1mm}$, which is adequate for the target application. A camera is also mounted to the gripper providing a top view, this is used to record/study the interaction of the gripper hairy pads with the object being grasped, and as a means to judge the grasp quality in the experimental results. It could potentially provide feedback information regarding grasp stability, and object slippage. The gripper has an overall dimensions of $150\times100\times252mm$ and a total weight of $640$ grams. It operates in either position or force control modes. The former controls the gripper opening gap, while the latter closes the fingers until the commanded force is achieved, measured via calibrated actuator current.

\begin{figure*}[t!]
\centering
\includegraphics[width=1.0\textwidth]{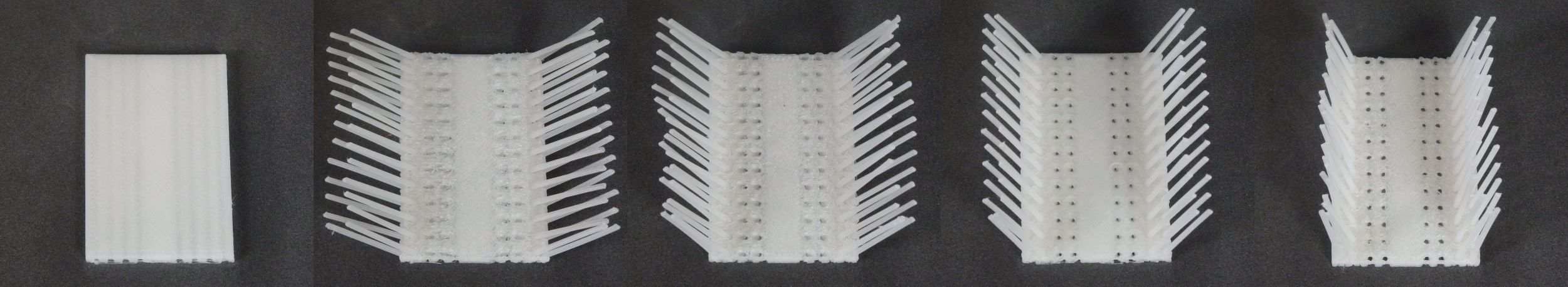}
\caption{Gripper pads used in evaluating the impact of hair angle on grasp success/robustness, showing (left to right) the hairless pad, hairy pads at $30\degree$, $45\degree$, $60\degree$, $75\degree$, all with hair density $D1$. These are compared in experiment 1, with corresponding results provided in Table \ref{table_grasping_results_varying_hair_angle}.}
\label{figure_set_of_gripper_pads_exp1}
\end{figure*}

\begin{figure}[b!]
\centering
\includegraphics[width=1.0\linewidth]{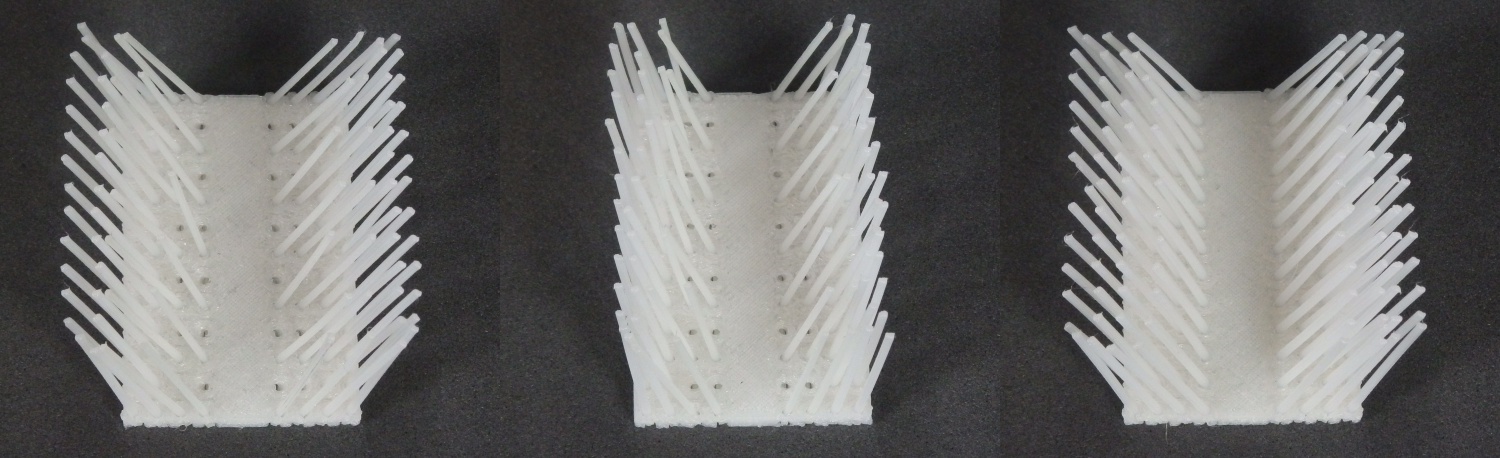}
\captionsetup{belowskip=-10pt}
\captionsetup{aboveskip=5pt}
\caption{Gripper pads with altered densities: at $60\degree$-$D2$, $75\degree$-$D2$, and $60\degree$-$D3$ respectively from left to right. These are compared in experiment 2, and corresponding results are provided in Table \ref{table_grasping_results_varying_hair_density}.}
\label{figure_set_of_gripper_pads_exp2}
\end{figure}

\begin{figure*}[t!]
\centering
\includegraphics[width=1.0\textwidth]{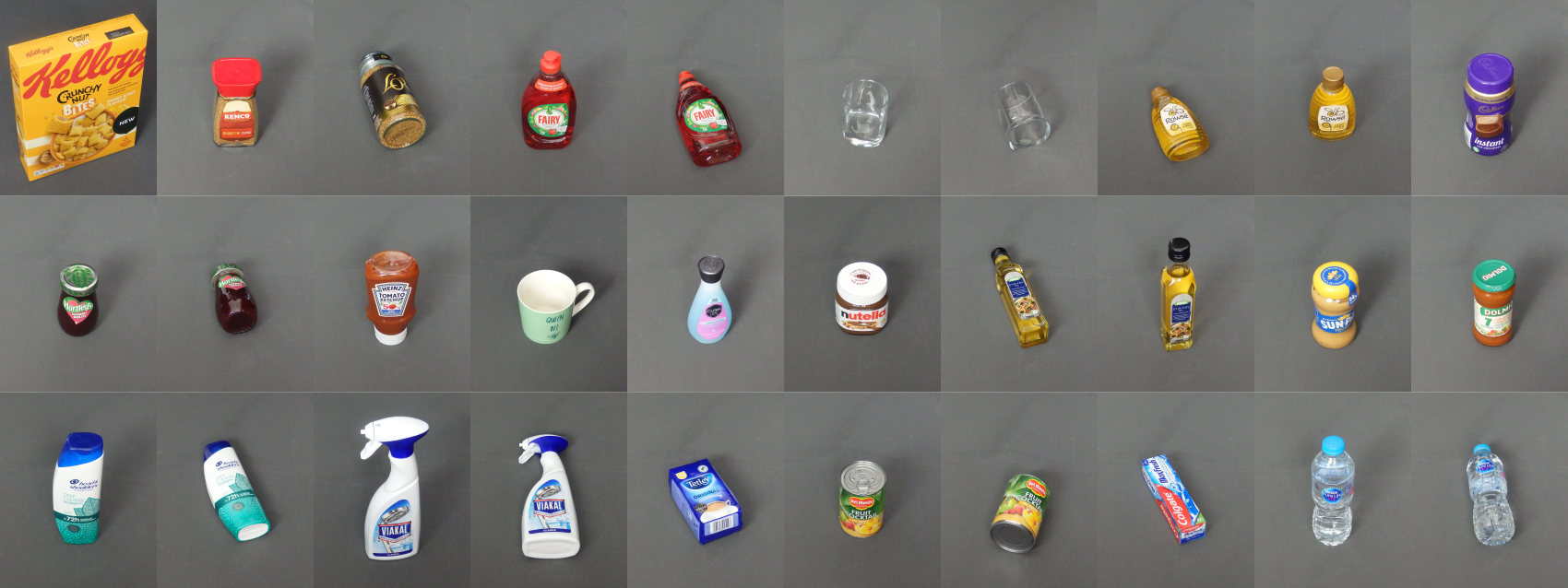}
\caption{Set of objects used to evaluate the proposed gripper.}
\label{figure_set_of_objects}
\end{figure*}

Two soft gripping pads are attached to each finger of the gripper. These can be fitted with hairs protruding at various angles/densities, or remain hairless. The pad shown in Fig. \ref{figure_gripper_CAD_model} has hairs at an angle of $60\degree$ with density pattern designated as $D1$ (see below). Fig. \ref{figure_pad_and_hair_CAD_model} shows the $60\degree$ gripping pad model in front, side, and back views respectively from left to right, and the two hair patterns used, with either $7$ or $13$ hairs, each hair is $20mm$ in length. The pad has a thickness of $2mm$ and contains a total of $104$ inclined holes, divided equally and symmetrically to the right and left sides of the pad's buffer (no hair) zone. All, some or none of these holes can be filled depending on the hair pattern used, and the hole inclination can be varied to alter the angle of the hairs. Fig. \ref{figure_set_of_gripper_pads_exp1} shows hair density pattern $D1$ made of $52$ hairs, for hair angles from $30-75\degree$. Fig. \ref{figure_set_of_gripper_pads_exp2} shows on the left (for two different angles) a density pattern $D2$ with $80$ hairs, representing a $54\%$ increase in hair density, and on the right, density pattern $D3$ with $104$ hairs, a $100\%$ hair density increase compared to $D1$. 

The gripping pads and hair patterns are made of thermoplastic polyurethane (TPU), with shore hardness $95$A via fused deposition modeling ($3D$ printed). This strikes a balance between firmness, so that the hairs can support grasped objects and retain their shape, and softness, so that the gripper can move seamlessly in tight spaces with the hairs deforming in compliance to allow objects inside the gripper with no disturbance. The fingers and pulleys are manufactured from Aluminium, the linear bearings and rail from steel, the remaining parts of polylactic acid (PLA).



\section{Experiments}
Two sets of experiments have been conducted, the first evaluates the effect of the hair angle on the grasp success at a given hair density $D1$ as compared to the no-hair performance. The gripper pads used in this experiment are shown in Fig. \ref{figure_set_of_gripper_pads_exp1}. The second experiment explores the effect of varying hair density for the best-performing hair angles, and the gripper pads used are depicted in Fig. \ref{figure_set_of_gripper_pads_exp2}. In the following subsections, the experiment design methodology and results are discussed.

\subsection{Setup and Methodology}
\begin{figure*}[t!]
\centering
\includegraphics[width=1.0\textwidth]{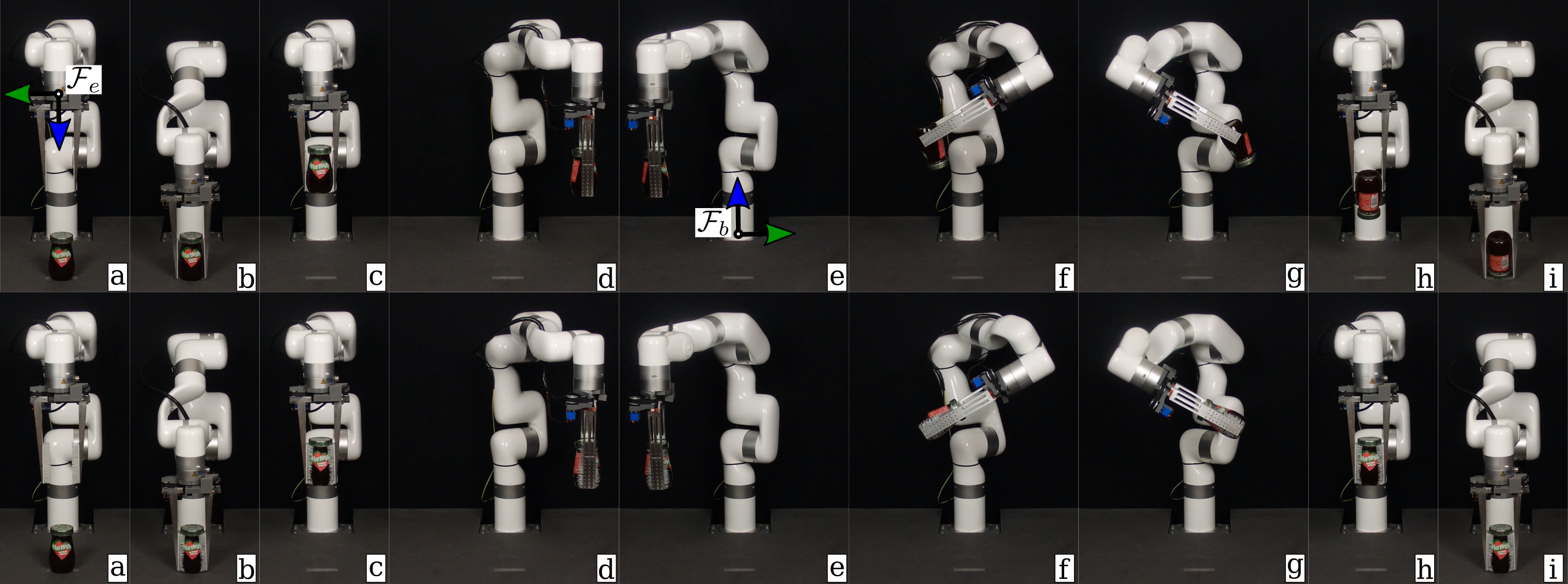}
\caption{Snapshots of the steps per experiment, starting with home pose, then grasping and lifting, manipulating to test the grasp stability, followed by stowing and releasing. Top is without hairs, bottom with hairs, producing better control of the object in poses f) and g).}
\label{figure_experiment_snapshots}
\end{figure*}

The developed gripper is fitted to the collaborative robot uFactory xArm$6$, as shown in Fig. \ref{figure_Antgrip_setup}, capable of manipulating a payload of $5$Kg. Both the robot arm and the gripper are controlled in \texttt{C++} by sending $6D$ pose, and position/force commands respectively. In each experiment, the object to be grasped is placed on a fixed mark on the table, representing the centroid point of the gripper workspace projected onto the table plane (on which the object is placed) at the arm home pose, shown in Fig. \ref{figure_experiment_snapshots}(a).
For each gripper pad variation ($8$ of these are tested here), $30$ objects were tested. For each object, $10$ grasping attempts were executed. In each of these $10$ attempts, the object position is varied about the aforementioned mark by up to $\pm{10mm}$, simulating potential object localization error. Simultaneously, object orientation about the vertical axis (perpendicular to the table plane) is varied widely across grasping attempts. A total of $2400$ grasping trials have been performed, all of which are video documented\footnote{\url{https://youtube.com/playlist?list=PL_VeUJXluS-uf0-TSOzl0bYHYz1oU_E5n&si=YzJ6sKtmOAyB3I7B}.}. The set of objects used consist of $21$ unique objects, of which, $9$ objects are tested in two different poses, namely; in a lying down or upright pose, making up a total of $30$ objects for testing purposes, shown in Fig. \ref{figure_set_of_objects}. In selecting the objects, care was taken to pick half of them to be of cylindrical/circular shape with varying radii, as such objects present a major challenge for parallel plate grippers. This provides potential to highlight any contribution of the hairs to grasping success. The other half are objects with flat, symmetrical sides, usually easy to grasp by a parallel plate, to assess if the added hairs might diminish conventional gripper capabilities. The objects were otherwise selected to provide variety in texture and deformability. They weigh from $117$ to $571$ grams, with $410$ grams on average, high enough to excite the object dynamics during the manipulation phase.

Snapshots of the robot movement sequence in each experiment are presented in Fig. \ref{figure_experiment_snapshots}, comparing the conventional parallel plate gripper (top) to that with hair pads (bottom). Each grasp trial consists of $11$ consecutive steps, made up of $9$ arm movements and $2$ gripper actions, starting with the arm in the home pose $\textbf{\texttt{P}}_\texttt{{HOME}}$, Fig. \ref{figure_experiment_snapshots}(a), and the gripper in a predefined opening gap depending on the width of the object to be grasped. The first three steps attempt to grasp and lift the object:
\begin{enumerate}
    \item The arm descends to the pose $\textbf{\texttt{P}}_\texttt{{GRASP}}$
    \item The gripper closes in force control mode until the command force is achieved (ranging from $10$ to $40N$ depending on the object's damage-resistance), Fig. \ref{figure_experiment_snapshots}(b).
    \item The arm then lifts back to the $\textbf{\texttt{P}}_\texttt{{HOME}}$, Fig. \ref{figure_experiment_snapshots}(c).
 \end{enumerate} 
 To test the grasp firmness and stability, the arm then manipulates the grasped object into $4$ successive poses, designed to excite movement in a direction parallel to the plane of grasp friction simulating a worst case scenario:
 \begin{enumerate}
  \setcounter{enumi}{3}
    \item Pure translation to the right, ending in pose 
    $\textbf{\texttt{P}}_\texttt{{LIN1}}$ Fig. \ref{figure_experiment_snapshots}(d)
    \item Pure translation to the left, ending in pose
    $\textbf{\texttt{P}}_\texttt{{LIN2}}$ Fig. \ref{figure_experiment_snapshots}(e)
    \item A combination of translatory and rotary motion ending in pose $\textbf{\texttt{P}}_\texttt{{ROT1}}$, Fig. \ref{figure_experiment_snapshots}(f).
    \item The opposite combination of translatory and rotary motion ending in pose $\textbf{\texttt{P}}_\texttt{{ROT2}}$, Fig. \ref{figure_experiment_snapshots}(g).
\end{enumerate}
The final steps deposit and release the object (assuming it is still grasped): 
 \begin{enumerate}
   \setcounter{enumi}{7}
 \item The arm moves back to the $\textbf{\texttt{P}}_\texttt{{HOME}}$, Fig. \ref{figure_experiment_snapshots}(h)
    \item The arm descends to the $\textbf{\texttt{P}}_\texttt{{RELEASE}}$ (here identical to the $\textbf{\texttt{P}}_\texttt{{GRASP}}$), Fig. \ref{figure_experiment_snapshots}(i)
    \item The gripper opens back to the initial gap distance in position control mode.
    \item Finally the arm terminates at the $\textbf{\texttt{P}}_\texttt{{HOME}}$.
\end{enumerate}
The arm's built-in motion profile parameters are set to a maximum velocity, and acceleration of $300mm/s$, and $700mm/s^2$ respectively. Let the pose vector $\textbf{\texttt{P}} = \begin{bmatrix*} ^{b}\textbf{r}_{e}^{\top} & \!\!\! ^{b}\bm{\theta}_{e}^{\top} \end{bmatrix*}^{\top} \!\!\! \in \mathbb{R}^{6}$ of the manipulator's end effector frame $\bm{\mathcal{F}}_{e}$ expressed in the base frame $\bm{\mathcal{F}}_b$, shown in Fig. \ref{figure_experiment_snapshots}(a)(top) and Fig. \ref{figure_experiment_snapshots}(e)(top) respectively, define the task space coordinates. With $^{b}\textbf{r}_{e} = \begin{bmatrix*} x & y & z \end{bmatrix*}^{\top}$ being the position vector, and $^{b}\bm{\theta}_{e} = \begin{bmatrix*} \alpha & \beta & \gamma \end{bmatrix*}^{\top}$ denoting a minimal representation of orientation (\textit{roll, pitch, yaw} RPY variation of Euler angles), the aforementioned poses are given by:
\begin{flalign*}
\notag &\hspace{25pt} \textbf{\texttt{P}}_\texttt{{HOME}} = \begin{bmatrix*} 0.37 & 0 & 0.5 & \pi & 0 & 0 \end{bmatrix*}^{\top},\\ 
\notag &\hspace{25pt} \textbf{\texttt{P}}_\texttt{{GRASP}} = \begin{bmatrix*} 0.37 & 0 & 0.265 \sim 0.34 & \pi & 0 & 0 \end{bmatrix*}^{\top},\\
\notag &\hspace{25pt} \textbf{\texttt{P}}_\texttt{{LIN1}} = \begin{bmatrix*} 0.4 & 0.2 & 0.5 & \pi & 0 & \frac{\pi}{2} \end{bmatrix*}^{\top},\\ 
\notag &\hspace{25pt} \textbf{\texttt{P}}_\texttt{{LIN2}} = \begin{bmatrix*} 0.4 & -0.2 & 0.5 & \pi & 0 & \frac{\pi}{2} \end{bmatrix*}^{\top},\\ 
\notag &\hspace{25pt} \textbf{\texttt{P}}_\texttt{{ROT1}} = \begin{bmatrix*} 0.3 & 0.1 & 0.5 & \pi & \frac{\pi}{3} & \frac{\pi}{2} \end{bmatrix*}^{\top},\\ 
\notag &\hspace{25pt} \textbf{\texttt{P}}_\texttt{{ROT2}} = \begin{bmatrix*} 0.3 & -0.1 & 0.5 & \pi & -\frac{\pi}{3} & \frac{\pi}{2} \end{bmatrix*}^{\top},
\end{flalign*}
where the position, and orientation values presented in meters, and radians respectively. Note the $z$ coordinate value of $\textbf{\texttt{P}}_\texttt{{GRASP}}$ ranges from $0.265m$ to $0.34m$ depending on the height of the object to be grasped.

\setlength{\tabcolsep}{0.6em} 
\begin{table*}[b!]
\captionsetup{aboveskip=-1pt,belowskip=0pt}
\caption{Successful grasp trials (out of 10 attempts) for different objects with varying hair angles at fixed hair density $D1$. Best performing results designated as quality grasps, shown in bold, defined as those with the ordered pair $\texttt{(loose,firm)} \in \{ \texttt{(2,8)}, \texttt{(1,9)}, \texttt{(0,10)} \} $.}
\label{table_grasping_results_varying_hair_angle}
\begin{center}

\begin{tabular}{|l l|c|c||c|c||c|c||c|c||c|c||c|c|}
\hline
\multicolumn{2}{|c|}{\multirow[c]{2}{*}{Object}} & Weight & Grip & \multicolumn{2}{c||}{No hair} & \multicolumn{2}{c||}{Hair at $30\degree$ $D1$}  & \multicolumn{2}{c||}{Hair at $45\degree$ $D1$}  & \multicolumn{2}{c||}{Hair at $60\degree$ $D1$}  & \multicolumn{2}{c|}{Hair at $75\degree$ $D1$} \\
& &(grams)& force(N) & loose & firm & loose & firm & loose & firm & loose & firm & loose & firm\\
\hline
\hline
1&Cereals box             & $441$ & $30$      & \textcolor{black}{$\textbf{0}$} & \textcolor{black}{$\textbf{10}$} & \textcolor{black}{$\textbf{0}$} & \textcolor{black}{$\textbf{10}$} & \textcolor{black}{$\textbf{0}$} & \textcolor{black}{$\textbf{10}$} & \textcolor{black}{$\textbf{0}$} & \textcolor{black}{$\textbf{10}$} & \textcolor{black}{$\textbf{0}$} & \textcolor{black}{$\textbf{10}$}\\
\rowcolor{lightgray}
2&Coffee jar1             & $365$ & $40$      & $2$ & $5$  & $2$ & $6$  & $1$ & $8$  & \textcolor{black}{$\textbf{0}$} & \textcolor{black}{$\textbf{10}$} & \textcolor{black}{$\textbf{0}$} & \textcolor{black}{$\textbf{10}$}\\
3&Coffee jar2             & $542$ & $40$      & $1$ & $8$  & \textcolor{black}{$\textbf{0}$} & \textcolor{black}{$\textbf{10}$} & \textcolor{black}{$\textbf{0}$} & \textcolor{black}{$\textbf{10}$} & \textcolor{black}{$\textbf{0}$} & \textcolor{black}{$\textbf{10}$} & \textcolor{black}{$\textbf{0}$} & \textcolor{black}{$\textbf{10}$}\\
\rowcolor{lightgray}
4&Dishwash liquid pose1   & $422$ & $40$      & $0$ & $0$  & $0$ & $0$  & $0$ & $0$  & $0$ & $0$  & $0$ & $3$\\
5&Dishwash liquid pose2   & $422$ & $40$      & $0$ & $7$  & \textcolor{black}{$\textbf{0}$} & \textcolor{black}{$\textbf{10}$} & \textcolor{black}{$\textbf{0}$} & \textcolor{black}{$\textbf{10}$} & \textcolor{black}{$\textbf{0}$} & \textcolor{black}{$\textbf{10}$} & \textcolor{black}{$\textbf{0}$} & \textcolor{black}{$\textbf{10}$}\\
\rowcolor{lightgray}
6&Glass cup pose1         & $252$ & $30$      & $5$ & $4$  & \textcolor{black}{$\textbf{0}$} & \textcolor{black}{$\textbf{10}$} & \textcolor{black}{$\textbf{0}$} & \textcolor{black}{$\textbf{10}$} & \textcolor{black}{$\textbf{0}$} & \textcolor{black}{$\textbf{10}$} & \textcolor{black}{$\textbf{0}$} & \textcolor{black}{$\textbf{10}$}\\
7&Glass cup pose2         & $252$ & $30$      & $6$ & $2$  & \textcolor{black}{$\textbf{0}$} & \textcolor{black}{$\textbf{10}$} & \textcolor{black}{$\textbf{0}$} & \textcolor{black}{$\textbf{10}$} & $0$ & $9$  & \textcolor{black}{$\textbf{0}$} & \textcolor{black}{$\textbf{10}$}\\
\rowcolor{lightgray}
8&Honey bottle pose1      & $365$ & $40$      & $0$ & $9$  & $\textbf{1}$ & $\textbf{9}$  & $0$ & $9$  & \textcolor{black}{$\textbf{0}$} & \textcolor{black}{$\textbf{10}$} & $\textbf{1}$ & $\textbf{9}$\\
9&Honey bottle pose2      & $365$ & $40$      & $0$ & $5$  & $0$ & $7$  & $0$ & $6$  & $0$ & $7$  & $0$ & $7$\\
\rowcolor{lightgray}
10&Chocolate powder       & $491$ & $40$      & $0$ & $9$  & \textcolor{black}{$\textbf{0}$} & \textcolor{black}{$\textbf{10}$} & \textcolor{black}{$\textbf{0}$} & \textcolor{black}{$\textbf{10}$} & \textcolor{black}{$\textbf{0}$} & \textcolor{black}{$\textbf{10}$} & $\textbf{1}$ & $\textbf{9}$\\
11&Jam jar pose1          & $492$ & $40$      & $0$ & $0$  & $10$ & $0$ & $9$ & $0$  & $5$ & $5$  & $6$ & $4$\\
\rowcolor{lightgray}
12&Jam jar pose2          & $492$ & $40$      & $0$ & $3$  & $2$ & $4$  & $0$ & $6$  & $0$ & $6$  & $0$ & $5$\\
13&Ketchup bottle         & $478$ & $40$      & $0$ & $4$  & $8$ & $2$  & $1$ & $8$  & \textcolor{black}{$\textbf{1}$} & \textcolor{black}{$\textbf{9}$}  & $5$ & $5$\\
\rowcolor{lightgray}
14&Mug                    & $323$ & $40$      & $8$ & $0$  & $10$ & $0$ & $10$ & $0$ & $10$ & $0$ & $10$ & $0$\\
15&Nail polish            & $203$ & $30$      & $3$ & $0$  & $3$ & $0$  & $1$ & $3$  & $5$ & $4$  & \textcolor{black}{$\textbf{2}$} & \textcolor{black}{$\textbf{8}$}\\
\rowcolor{lightgray}
16&Nutella bottle         & $568$ & $40$      & $6$ & $1$  & $9$ & $1$  & $5$ & $5$  & $6$ & $4$  & $6$ & $4$\\
17&Oil bottle pose1       & $491$ & $40$      & $2$ & $2$  & \textcolor{black}{$\textbf{0}$} & \textcolor{black}{$\textbf{10}$} & $1$ & $6$  & \textcolor{black}{$\textbf{0}$} & \textcolor{black}{$\textbf{10}$} & $0$ & $8$\\
\rowcolor{lightgray}
18&Oil bottle pose2       & $491$ & $40$      & $7$ & $1$  & $\textbf{2}$ & $\textbf{8}$  & $2$ & $6$  & \textcolor{black}{$\textbf{1}$} & \textcolor{black}{$\textbf{9}$}  & \textcolor{black}{$\textbf{1}$} & \textcolor{black}{$\textbf{9}$}\\
19&Peanut butter bottle   & $448$ & $40$      & $8$ & $1$  & \textcolor{black}{$\textbf{0}$} & \textcolor{black}{$\textbf{10}$} & \textcolor{black}{$\textbf{0}$} & \textcolor{black}{$\textbf{10}$} & \textcolor{black}{$\textbf{0}$} & \textcolor{black}{$\textbf{10}$} & \textcolor{black}{$\textbf{0}$} & \textcolor{black}{$\textbf{10}$}\\
\rowcolor{lightgray}
20&Sauce bottle           & $562$ & $40$      & $2$ & $0$  & $0$ & $0$  & $0$ & $0$  & $4$ & $0$  & $8$ & $0$\\
21&Shampoo pose1          & $455$ & $40$      & \textcolor{black}{$\textbf{0}$} & \textcolor{black}{$\textbf{10}$} & \textcolor{black}{$\textbf{0}$} & \textcolor{black}{$\textbf{10}$} & \textcolor{black}{$\textbf{0}$} & \textcolor{black}{$\textbf{10}$} & \textcolor{black}{$\textbf{0}$} & \textcolor{black}{$\textbf{10}$} & \textcolor{black}{$\textbf{0}$} & \textcolor{black}{$\textbf{10}$}\\
\rowcolor{lightgray}
22&Shampoo pose2          & $455$ & $40$      & \textcolor{black}{$\textbf{0}$} & \textcolor{black}{$\textbf{10}$} & \textcolor{black}{$\textbf{0}$} & \textcolor{black}{$\textbf{10}$} & \textcolor{black}{$\textbf{0}$} & \textcolor{black}{$\textbf{10}$} & \textcolor{black}{$\textbf{0}$} & \textcolor{black}{$\textbf{10}$} & \textcolor{black}{$\textbf{0}$} & \textcolor{black}{$\textbf{10}$}\\
23&Sprayer pose1          & $571$ & $40$      & $3$ & $4$  & $0$ & $8$  & $1$ & $8$  & $1$ & $6$  & $3$ & $1$\\
\rowcolor{lightgray}
24&Sprayer pose2          & $571$ & $40$      & $0$ & $0$  & $0$ & $5$  & $1$ & $4$  & $0$ & $2$  & $0$ & $1$\\
25&Tea bag                & $138$ & $10$      & \textcolor{black}{$\textbf{0}$} & \textcolor{black}{$\textbf{10}$} & \textcolor{black}{$\textbf{0}$} & \textcolor{black}{$\textbf{10}$} & \textcolor{black}{$\textbf{0}$} & \textcolor{black}{$\textbf{10}$} & \textcolor{black}{$\textbf{0}$} & \textcolor{black}{$\textbf{10}$} & \textcolor{black}{$\textbf{0}$} & \textcolor{black}{$\textbf{10}$}\\
\rowcolor{lightgray}
26&Tin can pose1          & $484$ & $40$      & $8$ & $0$  & \textcolor{black}{$\textbf{0}$} & \textcolor{black}{$\textbf{10}$} & \textcolor{black}{$\textbf{0}$} & \textcolor{black}{$\textbf{10}$} & \textcolor{black}{$\textbf{0}$} & \textcolor{black}{$\textbf{10}$} & \textcolor{black}{$\textbf{0}$} & \textcolor{black}{$\textbf{10}$}\\
27&Tin can pose2          & $484$ & $40$      & $0$ & $0$  & $1$ & $7$  & $0$ & $8$  & $\textbf{2}$ & $\textbf{8}$  & \textcolor{black}{$\textbf{0}$} & \textcolor{black}{$\textbf{10}$}\\
\rowcolor{lightgray}
28&Tooth paste            & $117$ & $10$      & \textcolor{black}{$\textbf{0}$} & \textcolor{black}{$\textbf{10}$} & \textcolor{black}{$\textbf{0}$} & \textcolor{black}{$\textbf{10}$} & \textcolor{black}{$\textbf{0}$} & \textcolor{black}{$\textbf{10}$} & \textcolor{black}{$\textbf{0}$} & \textcolor{black}{$\textbf{10}$} & \textcolor{black}{$\textbf{0}$} & \textcolor{black}{$\textbf{10}$}\\
29&Water bottle pose1     & $529$ & $30$      & \textcolor{black}{$\textbf{0}$} & \textcolor{black}{$\textbf{10}$} & \textcolor{black}{$\textbf{0}$} & \textcolor{black}{$\textbf{10}$} & \textcolor{black}{$\textbf{0}$} & \textcolor{black}{$\textbf{10}$} & \textcolor{black}{$\textbf{0}$} & \textcolor{black}{$\textbf{10}$} & \textcolor{black}{$\textbf{0}$} & \textcolor{black}{$\textbf{10}$}\\
\rowcolor{lightgray}
30&Water bottle pose2     & $529$ & $30$      & $0$ & $6$  & \textcolor{black}{$\textbf{0}$} & \textcolor{black}{$\textbf{10}$} & \textcolor{black}{$\textbf{0}$} & \textcolor{black}{$\textbf{10}$} & \textcolor{black}{$\textbf{0}$} & \textcolor{black}{$\textbf{10}$} & \textcolor{black}{$\textbf{0}$} & \textcolor{black}{$\textbf{10}$}\\

\hline
\hline
\multicolumn{4}{|l||}{Total (out of 300 trials per gripper variation)}    & $61$ & $131$  & $48$ & $207$ & $32$ & $217$ & $35$ & $229$ & $43$ & $223$\\
\hline
\hline
\multicolumn{4}{|l||}{Average grasp rate $\texttt{(loose+firm)/300}$}    & \multicolumn{2}{c||}{$64\%$}  & \multicolumn{2}{c||}{$85\%$} & \multicolumn{2}{c||}{$83\%$} & \multicolumn{2}{c||}{$87\%$} & \multicolumn{2}{c|}{$88.7\%$}\\
\hline
\hline
\multicolumn{4}{|l||}{Number of objects with quality grasps}    & \multicolumn{2}{c||}{$6/30$}  & \multicolumn{2}{c||}{$17/30$} & \multicolumn{2}{c||}{$14/30$} & \multicolumn{2}{c||}{$19/30$} & \multicolumn{2}{c|}{$19/30$}\\
\hline
\end{tabular}
\end{center}
\end{table*}

\subsection{Results}

The results of the first set of experiments are provided in Table \ref{table_grasping_results_varying_hair_angle}, showing the effect of hair angle at a fixed hair density. If the object slips out of grasp while being lifted or at any other subsequent phase, the trial is deemed a failed grasp. In general, the grasp success rate increases proportionally with the hair angle up to $60\degree$. To better interpret the results, we classify successful grasps into \textit{firm} and \textit{loose}. Once an object is grasped and lifted, if any point of the object is displaced by more than $1cm$ relative to the gripper during the manipulation phase, the grasp is labelled as \textit{loose}, irrespective of the motion type (rotary or linear), otherwise, it is a \textit{firm} grasp. This is due to the fact that a grasped object, while still in contact with (supported by) the table, is not in a stable grasp condition. For each gripper pad variation, the average grasp rate (AGR) is computed as a percentage of the successful grasps ($firm+loose$) over the total of $300$ grasping trials performed. We further define that a \textit{quality grasp} has occured for a particular object when, for a total of $10$ grasping attempts, all were successful with no more than $2$ \textit{loose}, i.e. ,satisfying as such the ordered pair:
\begin{equation*}
(loose,firm) \in \{ (2,8), (1,9), (0,10) \}.
\end{equation*}
Quality grasps are marked as bold in the respective table. 

The hairless pad variation has the lowest AGR value ($64\%$) and the highest number of \textit{loose} grasps, performing worse than any type of hairy pad. The observed boost in performance at density $D1$ ranges from $29\%$, computed as $(83-64)/64$ in case of $45\degree$ hair angle, to $38\%$ at $75\degree$ hairs. However, recognising that for many applications, loose grasps would be regarded as failures, we also assess reliability using the \textit{quality grasps} metric defined above which may be of more practical relevance. This shows an even more significant improvement in performance. In terms of \textit{quality grasps}, the performance boost ranges from $133\%$, computed as $(14-6)/6$ in case of $45\degree$ hair angle, to $216\%$ for $75\degree$ hairs.

\setlength{\tabcolsep}{0.42em} 
\begin{table}[b!]
\captionsetup{aboveskip=-1pt,belowskip=0pt}
\caption{Successful grasp trials (out of 10 attempts) for different objects with varying hair density.}
\label{table_grasping_results_varying_hair_density}
\begin{center}

\begin{tabular}{|l||c|c||c|c||c|c|}
\hline
\multirow[c]{2}{*}{ \hspace{2.5mm} Object} & \multicolumn{2}{c||}{Hair at $60\degree$ $D2$}  & \multicolumn{2}{c||}{Hair at $75\degree$ $D2$}  & \multicolumn{2}{c|}{Hair at $60\degree$ $D3$} \\
& loose & firm & loose & firm & loose & firm\\
\hline
\hline
Cereals box& $\textbf{0}$ & $\textbf{10}$      &$\textbf{0}$ & $\textbf{10}$       & $\textbf{0}$ & $\textbf{10}$\\
\rowcolor{lightgray}
Coffee jar1& $\textbf{0}$ & $\textbf{10}$       & $\textbf{0}$ & $\textbf{10}$     & $\textbf{0}$ & $\textbf{10}$\\
Coffee jar2& $\textbf{0}$ & $\textbf{10}$      & $\textbf{0}$ & $\textbf{10}$      & $\textbf{0}$ & $\textbf{10}$\\
\rowcolor{lightgray}
Dishwash p1 & $0$ & $0$       & $0$ & $0$        & $0$ & $0$\\
Dishwash p2& $\textbf{0}$ & $\textbf{10}$      & $\textbf{0}$ & $\textbf{10}$      & $\textbf{0}$ & $\textbf{10}$\\
\rowcolor{lightgray}
Glass cup p1& $\textbf{0}$ & $\textbf{10}$      & $\textbf{0}$ & $\textbf{10}$     & $\textbf{0}$ & $\textbf{10}$\\
Glass cup p2& $\textbf{0}$ & $\textbf{10}$      & $\textbf{0}$ & $\textbf{10}$     & $\textbf{0}$ & $\textbf{10}$\\
\rowcolor{lightgray}
Honey bot. p1& $\textbf{0}$ & $\textbf{10}$      & $\textbf{0}$ & $\textbf{10}$    & $\textbf{0}$ & $\textbf{10}$\\
Honey bot. p2& $0$ & $8$     & $0$ & $6$      & $0$ & $8$\\
\rowcolor{lightgray}
Choc. powder& $\textbf{0}$ & $\textbf{10}$    & $\textbf{0}$ & $\textbf{10}$      & $\textbf{0}$ & $\textbf{10}$\\
Jam jar p1& $\textbf{0}$ & $\textbf{10}$      & $\textbf{0}$ & $\textbf{10}$      & $\textbf{0}$ & $\textbf{10}$\\
\rowcolor{lightgray}
Jam jar p2& $\textbf{2}$ & $\textbf{8}$       & $0$ & $8$       & $0$ & $9$\\
Ketchup bot.& $\textbf{2}$ & $\textbf{8}$       & $1$ & $8$     & $\textbf{2}$ & $\textbf{8}$\\
\rowcolor{lightgray}
Mug& $5$ & $5$              & $\textbf{0}$ & $\textbf{10}$      & $3$ & $7$\\
Nail polish& $3$ & $6$       & $\textbf{0}$ & $\textbf{10}$     & $1$ & $7$\\
\rowcolor{lightgray}
Nutella bot.& $6$ & $4$       & $5$ & $5$     & $6$ & $4$\\
Oil bot. p1& $\textbf{0}$ & $\textbf{10}$       & $0$ & $6$     & $\textbf{0}$ & $\textbf{10}$\\
\rowcolor{lightgray}
Oil bot. p2& $\textbf{0}$ & $\textbf{10}$       & $\textbf{0}$ & $\textbf{10}$    & $\textbf{0}$ & $\textbf{10}$\\
Peanut bot.& $\textbf{0}$ & $\textbf{10}$      & $\textbf{0}$ & $\textbf{10}$     & $\textbf{0}$ & $\textbf{10}$\\
\rowcolor{lightgray}
Sauce bot.& $7$ & $0$       & $\textbf{0}$ & $\textbf{10}$      & $\textbf{0}$ & $\textbf{10}$\\
Shampoo p1& $\textbf{0}$ & $\textbf{10}$      & $\textbf{0}$ & $\textbf{10}$      & $\textbf{0}$ & $\textbf{10}$\\
\rowcolor{lightgray}
Shampoo p2& $\textbf{0}$ & $\textbf{10}$      & $\textbf{0}$ & $\textbf{10}$      & $\textbf{0}$ & $\textbf{10}$\\
Sprayer p1& $5$ & $0$       & $2$ & $0$       & $5$ & $0$\\
\rowcolor{lightgray}
Sprayer p2& $0$ & $0$       & $0$ & $2$       & $0$ & $0$\\
Tea bag& $\textbf{0}$ & $\textbf{10}$      & $\textbf{0}$ & $\textbf{10}$         & $\textbf{0}$ & $\textbf{10}$\\
\rowcolor{lightgray}
Tin can p1& $\textbf{0}$ & $\textbf{10}$      & $\textbf{0}$ & $\textbf{10}$      & $\textbf{0}$ & $\textbf{10}$\\
Tin can p2& $\textbf{0}$ & $\textbf{10}$       & $\textbf{0}$ & $\textbf{10}$     & $\textbf{0}$ & $\textbf{10}$\\
\rowcolor{lightgray}
Tooth paste& $\textbf{0}$ & $\textbf{10}$      & $\textbf{0}$ & $\textbf{10}$     & $\textbf{0}$ & $\textbf{10}$\\
Water bot. p1& $\textbf{0}$ & $\textbf{10}$      & $\textbf{0}$ & $\textbf{10}$   & $\textbf{0}$ & $\textbf{10}$\\
\rowcolor{lightgray}
Water bot. p2& $\textbf{0}$ & $\textbf{10}$      & $\textbf{0}$ & $\textbf{10}$   & $\textbf{0}$ & $\textbf{10}$\\

\hline
\hline
\multicolumn{1}{|l||}{Total} & $30$ & $239$ & $8$ & $255$ & $17$ & $253$\\
\hline
\hline
\multicolumn{1}{|l||}{Av. grasp rate}  & \multicolumn{2}{c||}{$89.7\%$} & \multicolumn{2}{c||}{$87.7\%$} & \multicolumn{2}{c|}{$90\%$}\\
\hline
\hline
\multicolumn{1}{|l||}{Quality grasps}  & \multicolumn{2}{c||}{$22/30$} & \multicolumn{2}{c||}{$22/30$} & \multicolumn{2}{c|}{$22/30$}\\
\hline
\end{tabular}
\end{center}
\end{table}

Results of the second set of experiments are provided in Table \ref{table_grasping_results_varying_hair_density}, investigating the effect of hair density on grasp success. Since the hair angles of $60\degree$ and $75\degree$ performed almost equally well in the first set of experiments, these are compared at the $53\%$ higher hair density $D2$. The best performing of out of theser (by a small margin, the hair angle $60\degree$), is then tested at the $100\%$ higher density $D3$. In general, higher hair density resulted in better grasping performance. The boost in AGR and in the \textit{quality grasps} reached $40\%$ and $266\%$ respectively for a $60\degree$ protruding hairs at $D3$ density when compared to the hairless counterpart, reflecting a notable increase in the number of \textit{firm} grasps across a range of objects. 

\begin{figure}[t!]
\centering
\includegraphics[width=0.4\textwidth]{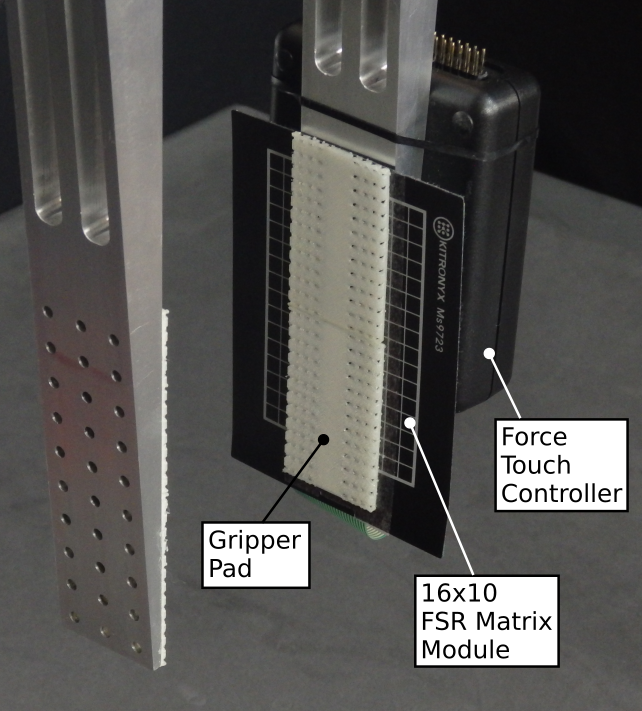}
\caption{Single-finger, hairless gripper-pad fitted with $16\times10$ FSR matrix.}
\label{figure_fsr_setup}
\end{figure}

\begin{figure}[t!]
\centering
\includegraphics[width=0.33\textwidth]{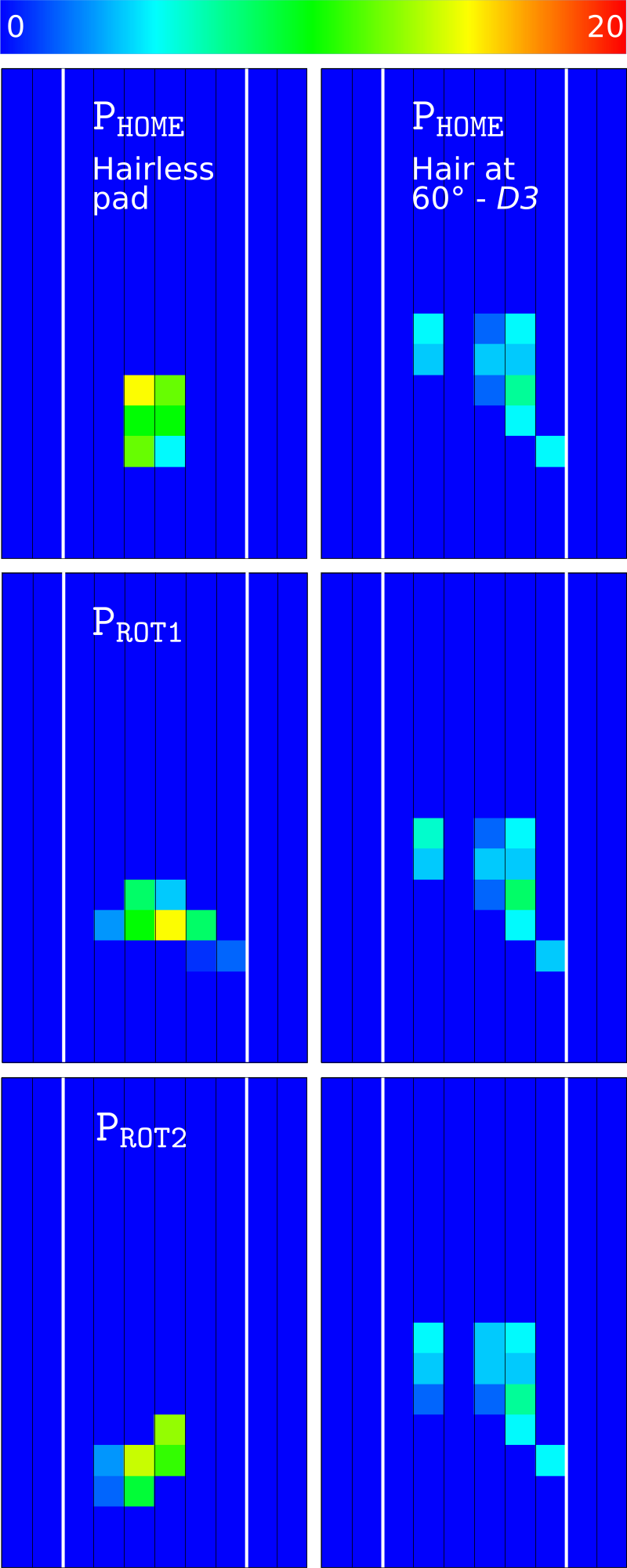}
\caption{Heat map of the $160$ FSR units showing the grasp force distribution for the hairy (right) and the hairless (left) pads. The two vertical white lines border the active area of the FSR matrix. Best performing hair pattern is used in comparison.}
\label{figure_fsr_result}
\end{figure}

\subsection{Grasp Force Distribution}
To better explain the improved grasping performance, a measure of the grasp force distribution is essential. To do so, the gripper finger setup is fitted with a force sensing resistor (FSR) matrix \cite{FSR_matrix} as shown in Fig. \ref{figure_fsr_setup} for the hairless pad. Due to the accuracy limitations of the FSR sensors, the objective of such setup is to compare the approximate grasp force distribution in the cases of hairy and hairless pads rather than an accurate force measurements. The result of such comparison is shown in Fig. \ref{figure_fsr_result}, featuring the \texttt{jam jar pose1} object, depicted in Fig. \ref{figure_experiment_snapshots}, with the corresponding force distribution heat maps of the hairy and hairless pads to the right and left columns respectively. The two vertical white lines of the respective figure border the active area of the FSR matrix that is in physical contact with the gripping pads. The heat map rows starting from top, show the force distribution at the experiment poses $\textbf{\texttt{P}}_\texttt{{HOME}}$, $\textbf{\texttt{P}}_\texttt{{ROT1}}$, and $\textbf{\texttt{P}}_\texttt{{ROT2}}$ respectively, these are depicted in Fig. \ref{figure_experiment_snapshots}(c), (f), and (g) in the same order.

From the results in Fig. \ref{figure_fsr_result} we can see that, the grasp force is better distributed in the case of the hairy pad, as opposed to larger force over a smaller area for the hairless pad. This essentially means the hairy pad applies lower pressure to the object, which can allow a higher grasp force for a more secure grasp. Also, we observe minimal change in the force distribution pattern for the hairy pad, indicating an almost stationary object, a sign of stable grasp. This is not the case with hairless pad, where the force pattern changes dramatically corresponding to an in-gripper moving object. 

\subsection{Discussion}

Biology has inspired a wide range of robot gripper designs, with recent examples including bird claws \cite{Roderick_2021}, bloodworms \cite{Dongbao2022}, geckos \cite{Ruotolo2021}, and many humanoid examples\cite{Nguyen2023}, as well as a wide variety of soft robotic end effectors. However it remains an issue that such designs typically introduce substantial complexity in construction, in control, or in both; and often for only limited manipulation improvement or advantage only in specialised scenarios. By considering a biological exemplar that more closely resembles the conventional two finger gripper, i.e., the mandibles of the ant, we have identified that their key function of securing grasped items during transport is aided by their hairy internal construction. This suggested a simple augmentation to a parallel jaw gripper --- the addition of hairs --- which we have shown here can provide a very significant improvement in handling of common household objects.

Looking more closely the results in Tables 1 and 2,
it is evident that some some objects are only grasped with particular hair angles, and this can differ for different objects. For example, the objects: \texttt{mug}, and \texttt{nail polish}, were only \textit{quality grasped} using the hair angle $75\degree$; and the \texttt{jam jar pose2}, \texttt{ketchup bottle}, and \texttt{oil bottle pose2} for the hair angle $60\degree$. The same conclusion applies to other hair angles in Table 1, where in the particular case of \texttt{sprayer pose1/2}, though never fully grasped, the hair angles $30\degree$ and $45\degree$ were the best performers. A general conclusion is that, the smaller the radius of curvature of the object's surface, the larger the hair angle required. This suggests that if a gripper with controllable hair angle, or a pad with several, but fixed hair angles is designed, it might be possible to combine the capabilities reported in Tables 1, 2, to produce 
 a \textit{quality grasps} total of $25$ for this particular set of objects. Note that the object \texttt{nutella bottle} enjoys a $100\%$ AGR only with hairy pads, but still produces no quality grasps; and it appears the cases of \texttt{dishwash liquid pose1}, \texttt{honey bottle pose2}, and \texttt{sprayer pose1/2} require additional friction to the pad/hair combination to be successful. Together with exploring means to achieve multiple hair angles this motivates future work to improve the gripper design.

In hairless pad experiments, during the rotary manipulation phase, objects usually pivot about pad points closer to the finger transmission, which suggests finger bending. Given the lengthy and slim fingers, there will always be some deflection in the gripper structure, specially at the fingers, irrespective of the quality and rigidity of the mechanical components, which can reduce but not eliminate its amplitude. Introducing internal hairs appears to overcome this issue by supporting the grasped object from the sides, increasing the area of contact and by doing so, better distributing the gripper's applied force on the object surface. The relatively hard material used in this work, TPU, is essential as such to firmly protrude out of the gripper to fill the gaps between the gripper and the object. During the experiments, the pads managed to return  close to their original shape fairly rapidly.

Given the level of grasp improvement we have already observed without these optimisations, we are confident that there should be many immediate applications in which AntGrip can improve performance. Looking forward we also plan to explore other aspects of the grasping capabilities of ants, such as their sensing and perceptual processing, for their potential to contribute to robotics.

\section{Conclusion}
In this work, a novel, ant inspired gripper is presented. The developed gripper has the parallel plate form,  suitable for tight space applications. Experimental evaluation on a pool of objects that are $50\%$ cylindrical in shape shows that adding flexible hairs to the interior of the plates not only improves the grasp success rate by at least $29\%$, but also the grasp robustness and quality. In addition, this result strongly suggests that the function of the internal hairs of ant mandibles goes beyond sensory feedback and plays an important mechanical role in their grasping ability.

\section*{ACKNOWLEDGMENTS}
This work is funded by EPSRC under grant agreement EP/V008102/1, An insect-inspired approach to robotic grasping.


\bibliographystyle{IEEEtran}  
\bibliography{main_tmech}

\begin{thebibliography}{10}
\providecommand{\url}[1]{#1}
\csname url@samestyle\endcsname
\providecommand{\newblock}{\relax}
\providecommand{\bibinfo}[2]{#2}
\providecommand{\BIBentrySTDinterwordspacing}{\spaceskip=0pt\relax}
\providecommand{\BIBentryALTinterwordstretchfactor}{4}
\providecommand{\BIBentryALTinterwordspacing}{\spaceskip=\fontdimen2\font plus
\BIBentryALTinterwordstretchfactor\fontdimen3\font minus \fontdimen4\font\relax}
\providecommand{\BIBforeignlanguage}[2]{{%
\expandafter\ifx\csname l@#1\endcsname\relax
\typeout{** WARNING: IEEEtran.bst: No hyphenation pattern has been}%
\typeout{** loaded for the language `#1'. Using the pattern for}%
\typeout{** the default language instead.}%
\else
\language=\csname l@#1\endcsname
\fi
#2}}
\providecommand{\BIBdecl}{\relax}
\BIBdecl

\bibitem{Correll2016}
N.~Correll, K.~E. Bekris, D.~Berenson, O.~Brock, A.~Causo, K.~Hauser, K.~Okada, A.~Rodriguez, J.~M. Romano, and P.~R. Wurman, ``Analysis and observations from the first amazon picking challenge,'' \emph{IEEE Transactions on Automation Science and Engineering}, vol.~15, no.~1, pp. 172--188, 2018.

\bibitem{Fujita2020}
M.~Fujita, Y.~Domae, A.~Noda, G.~A.~G. Ricardez, T.~Nagatani, A.~Zeng, S.~Song, A.~Rodriguez, A.~Causo, I.~M. Chen, and T.~Ogasawara, ``What are the important technologies for bin picking? technology analysis of robots in competitions based on a set of performance metrics,'' \emph{Advanced Robotics}, vol.~34, no. 7-8, pp. 560--574, 2020.

\bibitem{Sorour2019}
M.~{Sorour}, K.~{Elgeneidy}, A.~{Srinivasan}, M.~{Hanheide}, and G.~{Neumann}, ``Grasping unknown objects based on gripper workspace spheres,'' in \emph{2019 IEEE/RSJ International Conference on Intelligent Robots and Systems (IROS)}, Nov 2019, pp. 1541--1547.

\bibitem{Chen2015}
\BIBentryALTinterwordspacing
W.~Chen, S.~Zhao, and S.~L. Chow, \emph{Grippers and End-Effectors}.\hskip 1em plus 0.5em minus 0.4em\relax Springer London, 2015, pp. 2035--2070. [Online]. Available: \url{https://doi.org/10.1007/978-1-4471-4670-4\_96}
\BIBentrySTDinterwordspacing

\bibitem{Yu2016}
K.-T. Yu, N.~Fazeli, N.~Chavan-Dafle, O.~Taylor, E.~Donlon, G.~D. Lankenau, and A.~Rodriguez, ``A summary of team mit's approach to the amazon picking challenge 2015,'' 2016.

\bibitem{Gustavo2020}
R.~Gustavo, L.~El~Hafi, and F.~von Drigalski, ``Standing on giant’s shoulders: Newcomer’s experience from the amazon robotics challenge 2017,'' \emph{Advances on Robotic Item Picking: Applications in Warehousing and E-Commerce Fulfillment}, pp. 87--100, 2020.

\bibitem{Majumder2020}
\BIBentryALTinterwordspacing
A.~Majumder, O.~Kundu, S.~Dutta, S.~Kumar, and L.~Behera, ``Description of iitk-tcs system for arc 2017,'' \emph{Advances on Robotic Item Picking: Applications in Warehousing and E-Commerce Fulfillment}, pp. 113--124, 2020. [Online]. Available: \url{https://doi.org/10.1007/978-3-030-35679-8\_10}
\BIBentrySTDinterwordspacing

\bibitem{Morrison2017}
\BIBentryALTinterwordspacing
D.~Morrison, A.~W. Tow, M.~McTaggart, R.~Smith, N.~Kelly-Boxall, S.~Wade-McCue, J.~Erskine, R.~Grinover, A.~Gurman, T.~Hunn, D.~Lee, A.~Milan, T.~Pham, G.~Rallos, A.~Razjigaev, T.~Rowntree, K.~Vijay, Z.~Zhuang, C.~Lehnert, I.~Reid, P.~Corke, and J.~Leitner, ``Cartman: The low-cost cartesian manipulator that won the amazon robotics challenge,'' 2017. [Online]. Available: \url{https://arxiv.org/abs/1709.06283}
\BIBentrySTDinterwordspacing

\bibitem{Fujita2019}
M.~Fujita, Y.~Domae, R.~Kawanishi, G.~A.~G. Ricardez, K.~Kato, K.~Shiratsuchi, R.~Haraguchi, R.~Araki, H.~Fujiyoshi, S.~Akizuki, M.~Hashimoto, A.~Causo, A.~Noda, H.~Okuda, and T.~Ogasawara, ``Bin-picking robot using a multi-gripper switching strategy based on object sparseness,'' in \emph{2019 IEEE 15th International Conference on Automation Science and Engineering (CASE)}, 2019, pp. 1540--1547.

\bibitem{Hernandez2017}
C.~Hernandez, M.~Bharatheesha, W.~Ko, H.~Gaiser, J.~Tan, K.~van Deurzen, M.~de~Vries, B.~Van~Mil, J.~van Egmond, R.~Burger, M.~Morariu, J.~Ju, X.~Gerrmann, R.~Ensing, J.~Van~Frankenhuyzen, and M.~Wisse, ``Team delft's robot winner of the amazon picking challenge 2016,'' in \emph{RoboCup 2016: Robot World Cup XX}, S.~Behnke, R.~Sheh, S.~Sar{\i}el, and D.~D. Lee, Eds., 2017, pp. 613--624.

\bibitem{Huang2022}
H.~Huang, M.~Danielczuk, C.~M. Kim, L.~Fu, Z.~Tam, J.~Ichnowski, A.~Angelova, B.~Ichter, and K.~Goldberg, ``Mechanical search on shelves using a novel “bluction” tool,'' in \emph{2022 International Conference on Robotics and Automation (ICRA)}, 2022, pp. 6158--6164.

\bibitem{Valencia2017}
A.~J. Valencia, R.~M. Idrovo, A.~D. Sappa, D.~P. Guingla, and D.~Ochoa, ``A 3d vision based approach for optimal grasp of vacuum grippers,'' in \emph{2017 IEEE International Workshop of Electronics, Control, Measurement, Signals and their Application to Mechatronics (ECMSM)}, 2017.

\bibitem{Mantriota2011}
\BIBentryALTinterwordspacing
G.~Mantriota and A.~Messina, ``Theoretical and experimental study of the performance of flat suction cups in the presence of tangential loads,'' \emph{Mechanism and Machine Theory}, vol.~46, no.~5, pp. 607--617, 2011. [Online]. Available: \url{https://www.sciencedirect.com/science/article/pii/S0094114X11000139}
\BIBentrySTDinterwordspacing

\bibitem{Avigal2022}
\BIBentryALTinterwordspacing
Y.~Avigal, J.~Ichnowski, M.~Y. Cao, and K.~Goldberg, ``Gomp-st: Grasp optimized motion planning for suction transport,'' 2022. [Online]. Available: \url{https://arxiv.org/abs/2203.08359}
\BIBentrySTDinterwordspacing

\bibitem{Kang2019}
\BIBentryALTinterwordspacing
L.~Kang, J.-T. Seo, S.-H. Kim, W.-J. Kim, and B.-J. Yi, ``Design and implementation of a multi-function gripper for grasping general objects,'' \emph{Applied Sciences}, vol.~9, no.~24, 2019. [Online]. Available: \url{https://www.mdpi.com/2076-3417/9/24/5266}
\BIBentrySTDinterwordspacing

\bibitem{Pham2019}
H.~Pham and Q.-C. Pham, ``Critically fast pick-and-place with suction cups,'' in \emph{2019 International Conference on Robotics and Automation (ICRA)}, 2019, pp. 3045--3051.

\bibitem{Nakamoto2018}
H.~Nakamoto, M.~Ohtake, K.~Komoda, A.~Sugahara, and A.~Ogawa, ``A gripper system for robustly picking various objects placed densely by suction and pinching,'' in \emph{2018 IEEE/RSJ International Conference on Intelligent Robots and Systems (IROS)}, 2018, pp. 6093--6098.

\bibitem{WadeMcCue2017}
\BIBentryALTinterwordspacing
S.~Wade-McCue, N.~Kelly-Boxall, M.~McTaggart, D.~Morrison, A.~W. Tow, J.~Erskine, R.~Grinover, A.~Gurman, T.~Hunn, D.~Lee, A.~Milan, T.~Pham, G.~Rallos, A.~Razjigaev, T.~Rowntree, R.~Smith, K.~Vijay, Z.~Zhuang, C.~Lehnert, I.~Reid, P.~Corke, and J.~Leitner, ``Design of a multi-modal end-effector and grasping system: How integrated design helped win the amazon robotics challenge,'' 2017. [Online]. Available: \url{https://arxiv.org/abs/1710.01439}
\BIBentrySTDinterwordspacing

\bibitem{Nechyporenko2021}
\BIBentryALTinterwordspacing
N.~Nechyporenko, A.~Morales, E.~Cervera, and A.~P. del Pobil, ``A practical approach for picking items in an online shopping warehouse,'' \emph{Applied Sciences}, vol.~11, no.~13, 2021. [Online]. Available: \url{https://www.mdpi.com/2076-3417/11/13/5805}
\BIBentrySTDinterwordspacing

\bibitem{Wang2019}
S.~Wang, X.~Jiang, J.~Zhao, X.~Wang, W.~Zhou, and Y.~Liu, ``Vision based picking system for automatic express package dispatching,'' in \emph{2019 IEEE International Conference on Real-time Computing and Robotics (RCAR)}, 2019, pp. 797--802.

\bibitem{Hasegawa2017}
S.~Hasegawa, K.~Wada, Y.~Niitani, K.~Okada, and M.~Inaba, ``A three-fingered hand with a suction gripping system for picking various objects in cluttered narrow space,'' in \emph{2017 IEEE/RSJ International Conference on Intelligent Robots and Systems (IROS)}, 2017, pp. 1164--1171.

\bibitem{Hasegawa2019}
S.~Hasegawa, K.~Wada, K.~Okada, and M.~Inaba, ``A three-fingered hand with a suction gripping system for warehouse automation,'' \emph{Journal of Robotics and Mechatronics}, vol.~31, no.~2, pp. 289--304, 2019.

\bibitem{Vu2020}
Q.~Vu and A.~Ronzhin, ``A model of four-finger gripper with a built-in vacuum suction nozzle for harvesting tomatoes,'' in \emph{Proceedings of 14th International Conference on Electromechanics and Robotics}, 2020, pp. 149--160.

\bibitem{Wu2018}
P.-C. Wu, N.~Lin, T.~Lei, Q.~Cheng, J.-Z. Wu, and X.-P. Chen, ``A new grasping mode based on a sucked-type underactuated hand,'' \emph{Chinese Journal of Mechanical Engineering}, vol.~31, no.~1, 2018.

\bibitem{Yamaguchi2013}
K.~Yamaguchi, Y.~Hirata, and K.~Kosuge, ``Development of robot hand with suction mechanism for robust and dexterous grasping,'' in \emph{2013 IEEE/RSJ International Conference on Intelligent Robots and Systems}, 2013, pp. 5500--5505.

\bibitem{ant_photos}
\BIBentryALTinterwordspacing
K.~Stajniak, ``Antonyo photography,'' accessed: December 2022. [Online]. Available: \url{https://ant-photo.eu/en/}
\BIBentrySTDinterwordspacing

\bibitem{RightPick}
\BIBentryALTinterwordspacing
\text{RightHand Robotics Inc.}, ``Rightpick2,'' accessed: November 2022. [Online]. Available: \url{https://www.righthandrobotics.com/products}
\BIBentrySTDinterwordspacing

\bibitem{Wang2021}
\BIBentryALTinterwordspacing
Z.~Wang, H.~Furuta, S.~Hirai, and S.~Kawamura, ``A scooping-binding robotic gripper for handling various food products,'' \emph{Frontiers in Robotics and AI}, vol.~8, 2021. [Online]. Available: \url{https://www.frontiersin.org/articles/10.3389/frobt.2021.640805}
\BIBentrySTDinterwordspacing

\bibitem{Kobayashi2019}
A.~Kobayashi, J.~Kinugawa, S.~Arai, and K.~Kosuge, ``Design and development of compactly folding parallel open-close gripper with wide stroke,'' in \emph{2019 IEEE/RSJ International Conference on Intelligent Robots and Systems (IROS)}, 2019, pp. 2408--2414.

\bibitem{Guo2017}
M.~Guo, D.~V. Gealy, J.~Liang, J.~Mahler, A.~Goncalves, S.~McKinley, J.~A. Ojea, and K.~Goldberg, ``Design of parallel-jaw gripper tip surfaces for robust grasping,'' in \emph{2017 IEEE International Conference on Robotics and Automation (ICRA)}, 2017, pp. 2831--2838.

\bibitem{Harada2011}
K.~Harada, T.~Tsuji, K.~Nagata, N.~Yamanobe, K.~Maruyama, A.~Nakamura, and Y.~Kawai, ``Grasp planning for parallel grippers with flexibility on its grasping surface,'' in \emph{2011 IEEE International Conference on Robotics and Biomimetics}, 2011, pp. 1540--1546.

\bibitem{mason2018toward}
M.~T. Mason, ``Toward robotic manipulation,'' \emph{Annual Review of Control, Robotics, and Autonomous Systems}, vol.~1, no.~1, 2018.

\bibitem{Zhang2020}
W.~Zhang, Z.~He, Y.~Sun, J.~Wu, and Z.~Wu, ``A mathematical modeling method elucidating the integrated gripping performance of ant mandibles and bio-inspired grippers,'' \emph{Journal of Bionic Engineering}, vol.~17, no.~4, 2020.

\bibitem{Gronenberg1994}
\BIBentryALTinterwordspacing
W.~Gronenberg and J.~Tautz, ``The sensory basis for the trap-jaw mechanism in the ant odontomachus bauri,'' \emph{Journal of Comparative Physiology A}, vol. 174, pp. 49--60, 1994. [Online]. Available: \url{https://doi.org/10.1007/BF00192005}
\BIBentrySTDinterwordspacing

\bibitem{Boublil2021}
\BIBentryALTinterwordspacing
B.~L. Boublil, C.~A. Diebold, and C.~F. Moss, ``Mechanosensory hairs and hair-like structures in the animal kingdom: Specializations and shared functions serve to inspire technology applications,'' \emph{Sensors}, vol.~21, no.~19, 2021. [Online]. Available: \url{https://www.mdpi.com/1424-8220/21/19/6375}
\BIBentrySTDinterwordspacing

\bibitem{Wheeler1907}
W.~M. Wheeler, ``On certain modified hairs peculiar to the ants of arid regions,'' \emph{Biological Bulletin}, vol.~13, no.~4, pp. 185--202, 1907.

\bibitem{Spangler1966}
H.~G. Spangler and C.~W. Rettenmeyer, ``The function of the ammochaetae or psammophores of harvester ants, pogonomyrmex spp.'' \emph{Journal of the Kansas Entomological Society}, vol.~39, no.~4, pp. 739--745, 1966.

\bibitem{Porter1990}
S.~D. Porter and C.~D. Jorgensen, ``Psammophores: Do harvester ants (hymenoptera: Formicidae) use these pouches to transport seeds?'' \emph{Journal of the Kansas Entomological Society}, vol.~63, no.~1, pp. 138--149, 1990.

\bibitem{PAUL2001}
J.~Paul, ``Mandible movements in ants,'' \emph{Comparative Biochemistry and Physiology Part A: Molecular \& Integrative Physiology}, vol. 131, no.~1, pp. 7--20, 2001.

\bibitem{Klunk2021}
K.~Cristian, A.~Marco, C.-F. Alexandre, E.~Evan, and P.~Marcio, ``Mandibular morphology, task specialization and bite mechanics in pheidole ants (hymenoptera: Formicidae),'' \emph{Journal of the Royal Society}, vol. 18(179), 2021.

\bibitem{Gronenberg1997}
W.~Gronenberg, J.~Paul, S.~Just, and B.~Hölldobler, ``Mandible muscle fibers in ants: fast or powerful?'' \emph{Cell and Tissue Research}, vol. 289(2), 1997.

\bibitem{FSR_matrix}
kitronyx.com, ``Kitronyx fsr matrix and force controller,'' \url{https://www.kitronyx.com/store/p68/Tinn_Force_Touch_Controller.html}.

\bibitem{Roderick_2021}
\BIBentryALTinterwordspacing
W.~R.~T. Roderick, M.~R. Cutkosky, and D.~Lentink, ``Bird-inspired dynamic grasping and perching in arboreal environments,'' \emph{Science Robotics}, vol.~6, no.~61, p. eabj7562, 2021. [Online]. Available: \url{https://www.science.org/doi/abs/10.1126/scirobotics.abj7562}
\BIBentrySTDinterwordspacing

\bibitem{Dongbao2022}
\BIBentryALTinterwordspacing
D.~Sui, Y.~Zhu, S.~Zhao, T.~Wang, S.~K. Agrawal, H.~Zhang, and J.~Zhao, ``A bioinspired soft swallowing gripper for universal adaptable grasping,'' \emph{Soft Robotics}, vol.~9, no.~1, pp. 36--56, 2022. [Online]. Available: \url{https://doi.org/10.1089/soro.2019.0106}
\BIBentrySTDinterwordspacing

\bibitem{Ruotolo2021}
\BIBentryALTinterwordspacing
W.~Ruotolo, D.~Brouwer, and M.~R. Cutkosky, ``From grasping to manipulation with gecko-inspired adhesives on a multifinger gripper,'' \emph{Science Robotics}, vol.~6, no.~61, p. eabi9773, 2021. [Online]. Available: \url{https://www.science.org/doi/abs/10.1126/scirobotics.abi9773}
\BIBentrySTDinterwordspacing

\bibitem{Nguyen2023}
\BIBentryALTinterwordspacing
V.~P. Nguyen, S.~B. Dhyan, V.~Mai, B.~S. Han, and W.~T. Chow, ``Bioinspiration and biomimetic art in robotic grippers,'' \emph{Micromachines}, vol.~14, no.~9, 2023. [Online]. Available: \url{https://www.mdpi.com/2072-666X/14/9/1772}
\BIBentrySTDinterwordspacing

\end{thebibliography}

\newpage

\begin{IEEEbiography}[{\includegraphics[width=1in,height=1.25in,clip,keepaspectratio]{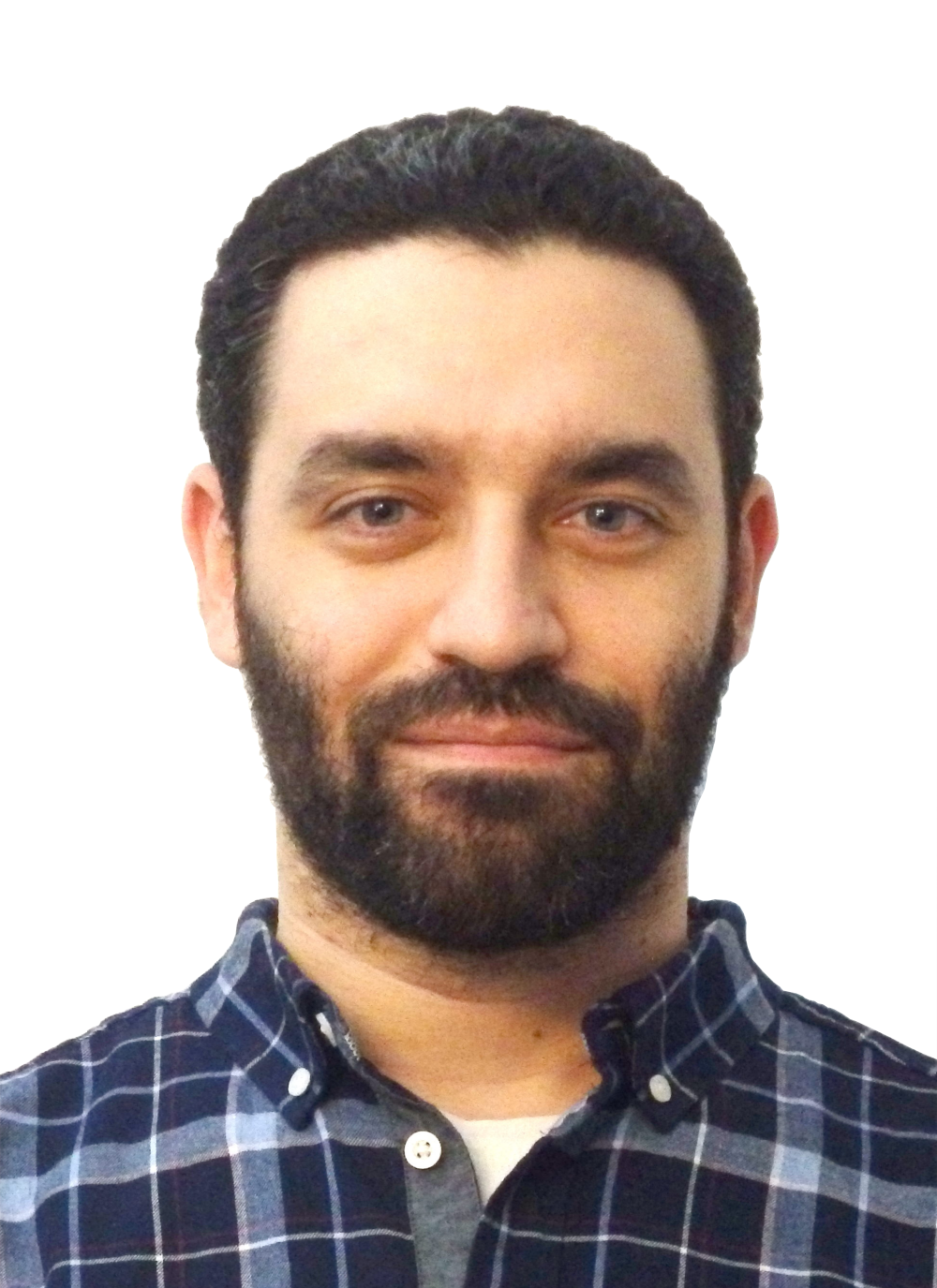}}]{Mohamed Sorour}
is a senior researcher at the University of Edinburgh. He finished his PhD in robotics at the University of Montpellier in 2017. Followed by postdoctoral research positions at the University of Lincoln and the Norwegian University of Life Sciences from 2018 to 2022 respectively. His research interests span bio-inspired robotic grasping, manipulation, mobile and agriculture robotics in addition to object segmentation.
\end{IEEEbiography}

\vspace{11pt}

\begin{IEEEbiography}[{\includegraphics[width=1in,height=1.25in,clip,keepaspectratio]{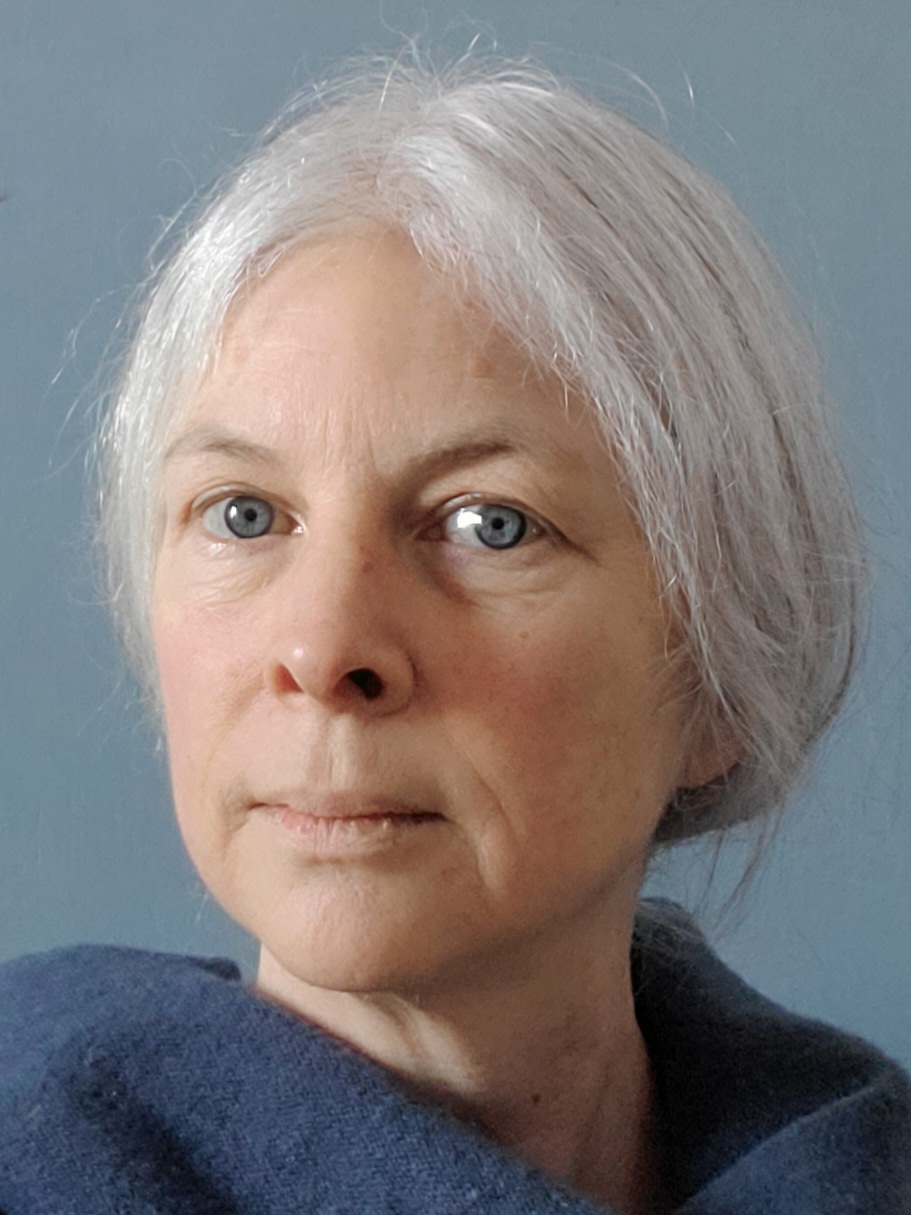}}]{Barbara Webb}
completed a BSc in Psychology at the University of Sydney
then a PhD in Artificial Intelligence at the University of Edinburgh,
during which she established a novel methodology of using embodied robot
models to evaluation biological hypotheses of behavioural control, with a
particular focus on insect sensorimotor systems. She has held lectureships
at the University of Nottingham and University of Stirling before
returning to a faculty position in the School of Informatics at Edinburgh
in 2003. She was appointed to a personal chair as Professor of Biorobotics
in 2010. Since that time the focus of her research has moved towards more
complex insect behavioural capabilities, such as learning and navigation,
implemented in neural models and tested on robot platforms. She was made a
fellow of the Royal Society of Edinburgh in 2022.
\end{IEEEbiography}

\vfill

\end{document}